\documentclass[12pt]{article}
\usepackage{framed,epsf,latexsym,amsmath,amssymb,amscd,textcomp}
\usepackage[numbers]{natbib}                                 
\usepackage{graphicx}
\usepackage[pdftex,colorlinks]{hyperref}
\usepackage[margin=1.0in]{geometry}

\newtheorem{theorem}{Theorem}[section]
\newtheorem{lemma}[theorem]{Lemma}
\newtheorem{remark}[theorem]{Remark}
\newtheorem{note}[theorem]{Note}
\newtheorem{example}[theorem]{Example}
\newtheorem{terminology}[theorem]{Terminology}
\newtheorem{counter-example}[theorem]{Counter example}
\newtheorem{proposition}[theorem]{Proposition}

\newtheorem{assumption}[theorem]{Assumption}
\newtheorem{open question}[theorem]{Open question}
\newtheorem{corollary}[theorem]{Corollary}
\newtheorem{conjecture}[theorem]{Conjecture}

\newtheorem{claim}{Claim}

\newcommand{\ignore}[1]{}

\newcommand{\ca}{{\cal A}}

\newcommand{\cd}{{\cal D}}

\newcommand{\cg}{{\cal G}}
\newcommand{\ch}{{\cal H}}

\newcommand{\cl}{{\cal L}}

\newcommand{\cx}{{\cal X}}

\newcommand{\cz}{{\cal Z}}

\DeclareMathOperator*{\sign}{sign}

\DeclareMathOperator*{\Lval}{\underline{VAL}}
\DeclareMathOperator*{\Uval}{\overline{VAL}}

\DeclareMathOperator*{\maj}{MAJ}

\newcommand{\csp}{\mathrm{CSP}}
\newcommand{\rsa}{\mathrm{RSA}}
\newcommand{\srcsp}{\mathrm{SRCSP}}
\newcommand{\auto}{\mathrm{AUTO}}
\newcommand{\rcsp}{\mathrm{RCSP}}
\newcommand{\res}{\mathrm{RES}}
\newcommand{\rand}{\mathrm{rand}}
\newcommand{\width}{\mathrm{width}}
\newcommand{\sat}{\mathrm{SAT}}
\newcommand{\dnf}{\mathrm{DNF}}

\newcommand{\var}{\mathrm{VAR}}
\newcommand{\val}{\mathrm{VAL}}
\newcommand{\half}{\mathrm{HALFSPACES}}
\newcommand{\parity}{\mathrm{PARITY}}
\newcommand{\inter}{\mathrm{INTER}}
\newcommand{\np}{\mathrm{NP}}

\newcommand{\proof}{{\par\noindent {\bf Proof}\space\space}}
\newcommand{\proofbox}{\hfill $\Box$}

\DeclareMathOperator{\Err}{Err}
\DeclareMathOperator{\poly}{poly}
\DeclareMathOperator*{\E}{\mathbb{E}}

\newcommand{\D}{\mathcal{D}}
\newcommand{\inner}[1]{\langle #1 \rangle}

\title{From average case complexity to improper learning complexity}

\author{Amit Daniely\thanks{Dept. of Mathematics, The Hebrew University, Jerusalem, Israel}   \hspace{1cm} Nati Linial\thanks{School of Computer Science and Engineering, The Hebrew University, Jerusalem, Israel.} \hspace{1cm}Shai Shalev-Shwartz\thanks{School of Computer Science and Engineering, The Hebrew University, Jerusalem, Israel} 
}

\begin{document}
\maketitle
\setcounter{page}{0}

\thispagestyle{empty}
\maketitle

\begin{abstract}
  The basic problem in the PAC model of computational learning theory is to determine
  which hypothesis classes are efficiently learnable. There is
  presently a dearth of results showing hardness of learning
  problems. Moreover, the existing lower bounds fall short of the best
  known algorithms.

The biggest challenge in proving complexity results is to establish hardness of {\em improper learning} (a.k.a. representation independent learning).
The difficulty in proving lower bounds for improper learning is that the standard reductions from $\mathbf{NP}$-hard problems do not seem to apply in this context.
There is essentially only one known approach to proving lower bounds on improper learning. It was initiated in \cite{KearnsVa89} and relies on cryptographic assumptions.

We introduce a new technique for proving hardness of improper learning, based on reductions from problems that are hard on average.
We put forward a (fairly strong) generalization of Feige's assumption
\citep{Feige02} about the complexity of refuting random constraint
satisfaction problems. Combining this assumption with our new technique yields far reaching implications. In particular,
\begin{itemize}
\item Learning $\mathrm{DNF}$'s is hard.
\item Agnostically learning halfspaces with a constant approximation ratio is hard.
\item Learning an intersection of $\omega(1)$ halfspaces is hard.
\end{itemize} 
\end{abstract}

\newpage

\section{Introduction}
Valiant's celebrated {\em probably approximately correct} (=PAC)
model~\citep{Valiant84} of machine learning led to an extensive
research that yielded a whole scientific community devoted to
computational learning theory. In the PAC learning model, a learner is
given an oracle access to randomly generated
samples $(X,Y)\in \cx\times\{0,1\}$ where $X$ is sampled from some
{\em unknown} distribution $\cd$ on $\cx$ and $Y=h^{*}(X)$ for some
{\em unknown} function $h^{*} : \cx \to \{0,1\}$. Furthermore, it is assumed that $h^*$
comes from a predefined \emph{hypothesis class} $\ch$, consisting of $0,1$ valued
functions on $\cx$. The learning problem defined by $\ch$ is to find a
function $h:\cx\to\{0,1\}$ that minimizes
$\Err_{\cd}(h):=\Pr_{X\sim\cd}(h(X)\not=h^*(X))$. For concreteness' sake we take
$\cx=\{\pm 1\}^n$, and we consider the learning problem tractable if there is an algorithm that on input
$\epsilon$, runs in time $\poly(n,1/\epsilon)$ and outputs, w.h.p., a
hypothesis $h$ with $\Err(h)\le\epsilon$.

Assuming $\mathbf{P}\ne\mathbf{NP}$, the status of most basic {\em computational} problems is fairly well understood. In a sharp contrast, almost $30$ years after Valiant's paper, the status of most basic {\em learning} problems is still wide open -- there is a huge gap between the performance of the best known algorithms and hardness results:
\begin{itemize}
\item No known algorithms can learn depth $2$ circuits, i.e., $\dnf$
  formulas. In contrast, we can only rule out learning of circuits of
  depth $d$, for some unspecified constant $d$
  \citep{Kharitonov93}. This result is based on a relatively strong
  assumption (a certain subexponential lower bound on factoring Blum
  integers). Under more standard assumptions (RSA in secure), the best
  we can do is rule out learning of depth $\log n$ circuits \citep{KearnsVa89}.
\item It is possible to agnostically learn halfspaces (see section
  \ref{sec:learning_background} for a definition of agnostic learning)
  with an approximation ratio of $O\left(\frac{n}{\log n}\right)$. On
  the other hand, the best known lower bound only rules out exact agnostic learning (\cite{FeldmanGoKhPo06},
  based on \cite{KlivansSh06}, under the assumption that the $\tilde{O}\left(n^{1.5}\right)$ unique shortest vector problem is hard).
\item No known algorithm can learn intersections of $2$ halfspaces,
  whereas Klivans and Sherstov~\citep{KlivansSh06} only rule out learning intersections of polynomially many halfspaces (again assuming that $\tilde{O}\left(n^{1.5}\right)$-uSVP is hard).
\end{itemize}
The crux of the matter, leading to this state of affairs, has to do
with the learner's freedom to return {\em any}
hypothesis. A learner who may return hypotheses outside
the class $\ch$ is called an
{\em improper learner}. This additional freedom
makes such algorithms potentially more powerful
than proper learners. On the other hand, this added flexibility makes it difficult to apply standard reductions from
$\mathbf{NP}$-hard problems. Indeed, there was no success so far in
proving intractability of a learning problem based on
$\mathbf{NP}$-hardness. Moreover, as Applebaum, Barak and Xiao
\cite{ApplebaumBaXi08} showed, many standard ways to do so are doomed
to fail, unless the polynomial hierarchy collapses.

The vast majority of existing lower bounds on learning utilize the
crypto-based argument, suggested in \cite{KearnsVa89}. Roughly
speaking, to prove that a certain learning problem is hard, one starts
with a certain collection of functions, that by assumption are one-way
trapdoor permutations. This immediately yields some hard (usually
artificial) learning problem. The final step is to reduce this
artificial problem to some natural learning problem.

Unlike the difficulty in establishing lower bounds for improper
learning, the situation in \emph{proper} learning is much better
understood. Usually, hardness of proper learning is
proved by showing that it is $\mathbf{NP}$-hard to distinguish a realizable sample from an unrealizable sample. I.e., it is hard to tell whether there is some
hypothesis in $\ch$ which has zero error on a given sample. This,
however, does not suffice for the purpose of proving lower bounds on
improper learning, because it might be the case that the learner finds a hypothesis (not from $\ch$) that does not err on the sample even though no $h\in\ch$ can accomplish this. In this paper we present a new
methodology for proving hardness of improper learning. Loosely
speaking, we show that improper learning is impossible provided that
it is hard to distinguish a realizable sample from a
\emph{randomly generated} unrealizable sample.

Feige \cite{Feige02} conjectured that random 3-$\sat$ formulas are hard to refute. He derived from this assumption certain hardness of approximation results, which are not known to follow from $\mathbf{P}\ne\mathbf{NP}$. We put forward a (fairly strong) assumption, generalizing Feige's assumption to certain predicates other that 3-$\sat$. Under this assumption, we show:
\begin{enumerate}
\item\label{echad} Learning $\mathrm{DNF}$'s is hard.
\item\label{shtaim} Agnostically learning halfspaces with a constant approximation ratio is hard, even over the boolean cube. 
\item\label{shalosh} Learning intersection of $\omega(1)$ halfspaces is hard, even over the boolean cube.
\item\label{arba} Learning finite automata is hard.
\item\label{chamesh} Learning parity is hard.
\end{enumerate}
We note that result~\ref{arba} can be established using the
cryptographic technique \citep{KearnsVa89}. Result~\ref{chamesh} is
often taken as a hardness {\em assumption}.  We also conjecture that
under our generalization of Feige's assumption it is hard to learn
intersections of even constant number of halfspaces. We present a
possible approach to the case of four halfspaces.  To the best
of our knowledge, these results easily imply most existing lower
bounds for improper learning.

\subsection{Comparison to the cryptographic technique}

There is a crucial reversal of order that works in our favour. To lower bound improper learning, we actually need much less than what is needed in cryptography, where
a problem and a distribution on instances are appropriate if they fool {\em every algorithm}.
In contrast, here we are presented with {\em a concrete} learning algorithms
and we devise a problem and a distribution on instances that fail it.

Second, cryptographic assumptions are often about the hardness
of number theoretic problems. In contrast, the average case assumptions
presented here are about $\csp$ problems. The proximity between $\csp$ problems
and learning problems is crucial for our purposes:
Since distributions are very
sensitive to gadgets, reductions between average case problems are much more limited than
reductions between worst case problems.

\subsection{On the role of average case complexity}

A key question underlying the present study and several additional recent papers is what can be deduced from the
average case hardness of specific problems.
Hardness on average is crucial for cryptography, and the security of
almost all modern cryptographic systems hinges on the average hardness
of certain problems, often from number theory. As shown by Kearns
and Valiant \citep{KearnsVa89}, the very same hardness on average
assumptions can be used to prove hardness of improper $\mathrm{PAC}$
learning of some hypothesis classes.

Beyond these classic results,
several recent works, starting from Feige's seminal work
\citep{Feige02}, show that average case hardness assumptions lead to
dramatic consequences in complexity theory.  The main idea of
\citep{Feige02} is to consider two possible avenues for progress
beyond the classic uses of average hardness: (i) Derive hardness in
additional domains, (ii) Investigate the implications of
hardness-on-average of other problems. For example, what are the
implications of average hardness of $3$-$\sat$? What about other
$\csp$ problems?

Feige \citep{Feige02} and then \citep{Alekhnovich03,BarakKiSt13} show
that average case hardness of $\csp$ problems have surprising
implications in hardness of approximation, much beyond
the consequences of standard complexity assumptions, or even cryptographic assumptions. Recently,
\citep{berthet2013computational} and \cite{daniely2013more} show that
hardness on average of planted clique and $3$-$\sat$ have implications
in learning theory, in the specific context of computational-sample
tradeoffs. In particular, they show that in certain learning tasks
(sparse $\mathrm{PCA}$ and learning halfspaces over sparse vectors)
more data can be leveraged to speed up computation. As we show here,
average case hardness of $\csp$ problems has implications even on the
hardness of very fundamental tasks in learning theory. Namely,
determining the tractability of $\mathrm{PAC}$ learning problems, most
of which are presently otherwise inaccessible.

\section{Preliminaries}
\subsection{Learning Theory}\label{sec:learning_background}
A {\em hypothesis class}, $\ch$, is a series of collections of functions $\ch_n\subset \{0,1\}^{\cx_n},\;n=1,2,\ldots$. We often abuse notation and identify $\ch$ with $\ch_n$. The instance space, $\cx_n$, that we consider is either $\cx_n=\{\pm 1\}^n$, $\cx_n=\{0, 1\}^n$ or $\cx_n=\{-1,1,0\}^n$. Concrete hypothesis classes, such as halfspaces, DNF's etc., are denoted $\half,\dnf$ etc. Also $\cz_n:=\cx_n\times\{0,1\}$.

Distributions on $\cz_n$ (resp. $\cz_n^m$) are denoted $\cd_n$ (resp. $\cd_n^m$). {\em Ensembles} of distributions are denoted by $\cd$. That is, $\cd=\{\cd_n^{m(n)}\}_{n=1}^\infty$ where $\cd_n^{m(n)}$ is a distributions on $\cz_n^{m(n)}$. We say that $\cd$ is a {\em polynomial ensemble} if $m(n)$ is upper bounded by some polynomial in $n$.

The error of a hypothesis $h:\cx_n\to\{0,1\}$ w.r.t. $\cd_n$ on $\cz_n$ is defined as $\Err_{\cd_n}(h)=\Pr_{(x,y)\sim\cd_n }\left(h(x)\ne y\right)$. For a hypothesis class $\ch_n$, we define $\Err_{\cd_n}(\ch_n)=\min_{h\in\ch_n}\Err_{\cd_n}(h)$. We say that a distribution
$\cd_n$ is \emph{realizable} by $h$ (resp. $\ch_n$) if $\Err_{\cd_n}(h)=0$ (resp. $\Err_{\cd_n}(\ch_n)=0$). Similarly, we say that $\cd_n$ is {\em $\epsilon$-almost realizable} by $h$ (resp. $\ch_n$)  if $\Err_{\cd_n}(h)\le\epsilon$ (resp. $\Err_{\cd_n}(\ch_n)\le\epsilon$).

A {\em sample} is a sequence $S=\{(x_1,y_1),\ldots (x_m,y_m)\}\in\cz^m_n$. The {\em empirical error} of a hypothesis $h:\cx_n\to\{0,1\}$ w.r.t. sample $S$ is $\Err_{S}(h)=\frac{1}{m}\sum_{i=1}^m1(h(x_i)=y_i)$. The {\em empirical error} of a hypothesis class $\ch_n$ w.r.t. $S$ is $\Err_{S}(\ch_n)=\min_{h\in\ch_n}\Err_S(h)$.
We say that a sample $S$ is \emph{realizable} by $h$ if $\Err_S(h)=0$. The sample $S$ is {\em realizable} by $\ch_n$ if $\Err_S(\ch_n)=0$. Similarly, we define the notion of {\em $\epsilon$-almost realizable} sample (by either a hypothesis $h:\cx_n\to\{0,1\}$ or a class $\ch_n$).

A {\em learning algorithm}, denoted $\cl$, obtains an error parameter $0<\epsilon<1$, a confidence parameter $0<\delta<1$, a complexity parameter $n$, and an access to an oracle that produces samples according to unknown distribution $\cd_n$ on $\cz_n$. It should output a (description of) hypothesis $h:\cx_n\to\{0,1\}$. We say that the algorithm $\cl$ {\em (PAC) learns} the hypothesis class $\ch$ if, for every realizable distribution $\cd_n$, with probability $\ge 1-\delta$, $\cl$ outputs a hypothesis with error $\le \epsilon$.
We say that an algorithm $\cl$ {\em agnostically learns} $\ch$ if, for every distribution $\cd_n$, with probability $\ge 1-\delta$, $\cl$ outputs a hypothesis with error $\le \Err_{\cd_n}(\ch)+\epsilon$. 
We say that an algorithm $\cl$ {\em approximately agnostically learns} $\ch$ with approximation ratio $\alpha=\alpha(n)\ge 1$ if, for every distribution $\cd_n$, with probability $\ge 1-\delta$, $\cl$ outputs a hypothesis with error $\le \alpha\cdot \Err_{\cd_n}(\ch)+\epsilon$. We say that $\cl$ is {\em efficient} if it runs in time polynomial in $n,1/\epsilon$ and $1/\delta$, and outputs a hypothesis that can be evaluated in time polynomial in $n,1/\epsilon$ and $1/\delta$. We say that $\cl$ is {\em proper} (with respect to $\ch$) if it always outputs a hypothesis in $\ch$. Otherwise, we say that $\cl$ is {\em improper}.

Let $\ch=\{\ch_n\subset\{0,1\}^{\cx_n}\mid n=1,2\ldots\}$ and $\ch'=\{\ch'_n\subset\{0,1\}^{\cx'_n}\mid n=1,2\ldots\}$ be two hypothesis classes. We say the $\ch$ is {\em realized} by $\ch'$ if there are functions $g:\mathbb N\to \mathbb N$ and $f_n:\cx_n\to\cx'_{g(n)},\;n=1,2,\ldots$ such that for every $n$, $\ch_n\subset \{h'\circ f_n\mid h'\in\ch'_n\}$. We say that $\ch$ is {\em efficiently realized} by $\ch'$ if, in addition, $f_n$ can be computed it time polynomial in $n$. Note that if $\ch'$ is efficiently learnable (respectively, agnostically learnable, or approximately agnostically learnable) and $\ch$ is efficiently realized by $\ch'$, then $\ch$ is efficiently learnable (respectively, agnostically learnable, or approximately agnostically learnable) as well.

\subsection{Constraints Satisfaction Problems}
Let $P:\{\pm 1\}^K\to \{0,1\}$ be some boolean predicate (that is, $P$
is any non-constant function from $\{\pm 1\}^K$ to $\{0,1\}$). A {\em
  $P$-constraint} with $n$ variables is a function $C:\{\pm
1\}^n\to\{0,1\}$ of the form $C(x)=P(j_1x_{i_1},\ldots,j_Kx_{i_K})$
for $j_l\in \{\pm 1\}$ and $K$ distinct $i_l\in [n]$. The {\em CSP
  problem,} $\csp(P)$, is the following. An instance to the problem is
a collection $J=\{C_1,\ldots,C_m\}$ of $P$-constraints and the
objective is to find an assignment $x\in \{\pm 1\}^n$ that maximizes
the fraction of satisfied constraints (i.e.,
constraints with $C_i(x)=1$). The {\em value} of the instance $J$,
denoted $\val(J)$, is the maximal fraction of constraints that can be
simultaneously satisfied. If $\val(J)=1$, we say that $J$ is
satisfiable.

For $1\ge\alpha>\beta>0$, the problem $\csp^{\alpha,\beta}(P)$ is the
decision promise problem of distinguishing between instances to
$\csp(P)$ with value $\ge \alpha$ and instances with value $\le
\beta$. Denote $\Lval(P)=\E_{x\sim \mathrm{Uni}(\{\pm 1\}^K)}P(x)$. We
note that for every instance $J$ to $\csp(P)$, $\val(J)\ge \Lval(P)$
(since a random assignment $\psi\in\{\pm 1\}^n$ satisfies in
expectation $\Lval(P)$ fraction of the constraints). Therefore, the
problem $\csp^{\alpha,\beta}(P)$ is non-trivial only if $\beta\ge
\Lval(P)$. We say that $P$ is {\em approximation resistant} if, for
every $\epsilon>0$, the problem
$\csp^{1-\epsilon,\Lval(P)+\epsilon}(P)$ is $\mathbf{NP}$-hard. Note
that in this case, unless $\mathbf{P}=\mathbf{NP}$, no algorithm for
$\csp(P)$ achieves better approximation ratio than the naive algorithm
that simply chooses a random assignment. We will use even stronger
notions of approximation resistance: We say that $P$ is {\em
  approximation resistant on satisfiable instances} if, for every
$\epsilon>0$, the problem $\csp^{1,\Lval(P)+\epsilon}(P)$ is
$\mathbf{NP}$-hard. Note that in this case, unless
$\mathbf{P}=\mathbf{NP}$, no algorithm for $\csp(P)$ achieves better
approximation ratio than a random assignment, even if the instance is
guaranteed to be satisfiable. We say that $P$ is {\em heredity
  approximation resistant on satisfiable instances} if every predicate
that is implied by $P$ (i.e., every predicate $P':\{\pm
1\}^K\to\{0,1\}$ that satisfies $\forall x,\;P(x)\Rightarrow P'(x)$)
is approximation resistant on satisfiable instances. Similarly, we
define the notion of {\em heredity approximation resistance}.

We will consider average case variant of the problem $\csp^{\alpha,\beta}(P)$. 
Fix $1\ge \alpha> \Lval(P)$. By a simple counting argument, for sufficiently large constant $C>0$, the value of a random instance with $\ge C\cdot n$ constraints is about $\Lval(P)$, in particular, the probability that a (uniformly) random instance to $\csp(P)$ with $n$ variables and $\ge Cn$ constraints will have value $\ge \alpha$ is exponentially small. Therefore, the problem of distinguishing between instances with value $\ge\alpha$ and random instances with $m(n)$ constraints can be thought as an average case analogue of $\csp^{\alpha,\Lval(P)+\epsilon}$. We denote this problem by $\csp^{\alpha,\rand}_{m(n)}(P)$. Precisely, we say that the problem $\csp^{\alpha,\rand}_{m(n)}(P)$ is easy, if there exists an efficient randomized algorithm, $\ca$, with the following properties:
\begin{itemize}
\item If $J$ is an instance to $\csp(P)$ with $n$ variables, $m(n)$ constraints, and value $\ge \alpha$, then
\[
\Pr_{\text{coins of }\ca}\left(\ca(J)=``\val(J)\ge\alpha"\right)\ge\frac{3}{4}
\]  
\item If $J$ is a random instance to $\csp(P)$ with $n$ variables and $m(n)$ constraints then, with probability $1-o_n(1)$ over the choice of $J$,
\[
\Pr_{\text{coins of }\ca}\left(\ca(J)=``J\text{ is random}"\right)\ge \frac{3}{4}~.
\]  
\end{itemize}
The problem $\csp^{\alpha,\rand}_{m(n)}(P)$ will play a central role. In particular, the case $\alpha=1$, that is, the problem of distinguishing between satisfiable instances and random instances. This problem is also known as the problem of refuting random instances to $\csp(P)$.
A simple observation is that the problem $\csp^{1,\rand}_{m(n)}(P)$ becomes easier as $m$ grows: If $m'\ge m$, we can reduce instances of $\csp^{1,\rand}_{m'(n)}(P)$ to instances of $\csp^{1,\rand}_{m(n)}(P)$ by simply drop the last $m'(n)-m(n)$ clauses. Note that if the original instance was either random or satisfiable, the new instance has the same property as well. Therefore, a natural metric to evaluate a refutation algorithm is the number of random constraints that are required to guarantee that the algorithm will refute the instance with high probability.

Another simple observation is that if a predicate $P':\{\pm 1\}^K\to
\{0,1\}$ is implied by $P$ then the problem
$\csp^{1,\rand}_{m(n)}(P')$ is harder than
$\csp^{1,\rand}_{m(n)}(P')$.  Indeed, given an instance to $\csp(P)$,
we can create an instance to $\csp(P')$ by replacing each constraint
$C(x)=P(j_1x_{i_1},\ldots,j_Kx_{j_K})$ with the constraint
$C'(x)=P'(j_1x_{i_1},\ldots,j_Kx_{j_K})$. We note that this reduction
preserves both satisfiability and randomness, and therefore
establishes a valid reduction from $\csp^{1,\rand}_{m(n)}(P)$ to
$\csp^{1,\rand}_{m(n)}(P')$.

\subsection{Resolution refutation and Davis Putnam algorithms}\label{sec:preliminaries_res}
A clause is a disjunction of literals, each of which correspond to a distinct variable. 
Given two clauses of the form $x_i\vee C$ and $\neg x_i\vee D$ for some clauses $C,D$, the {\em resolution rule} infer the clause $C\vee D$.
Fix a predicate $P:\{\pm 1\}^K\to \{0,1\}$. A {\em resolution refutation} for an instance $J=\{C_1,\ldots,C_m\}$ to $\csp(P)$ is a sequence of clauses $\tau=\{T_1,\ldots, T_r\}$ such that $T_r$ is the empty clause, and for every $1\le i\le r$, $T_i$ is either implied by some $C_j$ or resulted from the resolution rule applied on $T_{i_1}$ and $T_{i_2}$ for some $i_1,i_2<i$. We note that every un-satisfiable instance to $\csp(P)$ has a resolution refutation (of possibly exponential length). We denote by $\res(J)$ the length of the shortest resolution refutation of $J$.

The length of resolution refutation of random $K$-$\sat$ instances were extensively studied (e.g., \citep{BenWi99}, \citep{BeameKaPiSa98} and \citep{BeamePi96}). Two motivations for these study are the following. First, the famous result of \cite{CookRe79}, shows that $\mathbf{NP}\ne\mathbf{CoNP}$ if and only is there is no propositional proof system that can refute every instance $J$ to $K$-$\sat$ in length polynomial in $|J|$. Therefore, lower bound on concrete proof systems might bring us closer to $\mathbf{NP}\ne\mathbf{CoNP}$. Also, such lower bounds might indicate that refuting such instances in general, is intractable.

A second reason is that many popular algorithms implicitly produces a resolution refutation during their execution.
Therefore, any lower bound on the size of the resolution refutation would lead to the same lower bound on the running time or the algorithm.
A widely used and studied refutation algorithms of this kind are Davis-Putnam (DPLL) like algorithms \citep{davis1962machine}.  A DPLL algorithm is a form of recursive search for a satisfying assignment which on $\csp$ input $J$ operates
as follows: If $J$ contains the constant predicate $0$, it terminates and outputs that the instance is un-satisfiable. Otherwise, a variable $x_i$ is chosen, according to some rule. Each assignment to $x_i$ simplifies the instance $J$, and the algorithm recurses on these simpler instances.

\section{The methodology}\label{sec:methodology}
We begin by discussing the methodology in the realm of realizable learning, and we later proceed to agnostic learning. Some of the ideas underling our methodology appeared, in a much more limited context, in \cite{daniely2013more}.

To motivate the approach, recall how one usually proves that a class cannot be efficiently {\em properly} learnable. Given a hypothesis class $\ch$, let $\Pi(\ch)$ be the problem of distinguishing between an $\ch$-realizable sample  $S$ and one with $\Err_S(\ch)\ge \frac{1}{4}$. If $\ch$ is efficiently {\em properly} learnable then this problem is in\footnote{The reverse direction is almost true: If the search version of this problem can be solved in polynomial time, then $\ch$ is efficiently learnable.} $\mathbf{RP}$:
To solve $\Pi(\ch)$, we simply invoke a proper learning algorithm
$\ca$ that efficiently learns $\ch$, with examples drawn uniformly
from $S$. Let $h$ be the output of $\ca$. Since $\ca$ properly learns $\ch$, we have
\begin{itemize}
\item If $S$ is a realizable sample, then $\Err_S(h)$ is small.
\item If $\Err_S(\ch)\ge\frac{1}{4}$ then, {\em since $h\in\ch$}, $\Err_S(h)\ge \frac{1}{4}$.
\end{itemize}
This gives an efficient way to decide whether $S$ is realizable. We conclude that if $\Pi(\ch)$ is $\mathbf{NP}$-hard, then $\ch$ is not efficiently learnable, unless $\mathbf{NP}=\mathbf{RP}$.

However, this argument does not rule out the possibility that $\ch$ is
still learnable by an {\em improper} algorithm. Suppose now that $\ca$
efficiently and improperly learns $\ch$. If we try to use the above
argument to prove that $\Pi(\ch)$ can be efficiently solved, we get
stuck -- suppose that $S$ is a sample and we invoke $\ca$ on it, to
get a hypothesis $h$. As before, if $S$ is realizable, $\Err_S(h)$ is
small. However, if $S$ is not realizable, since $h$ not necessarily
belongs to $\ch$, it still might be the case that $\Err_S(h)$ is
small. Therefore, the argument fails.  We emphasize that this is not
only a mere weakness of the argument -- there are classes for which
$\Pi(\ch)$ is $\mathbf{NP}$-hard, but yet, they are learnable by an
improper algorithm\footnote{This is true, for example, for the class
  of $\dnf$ formulas with 3 $\dnf$ clauses.}. More generally,
Applebaum et al \cite{ApplebaumBaXi08} indicate that it is unlikely that hardness of
improper learning can be based on standard reductions from
$\mathbf{NP}$-hard problems, as the one described here.

We see that it is not clear how to establish hardness of improper
learning based on the hardness of distinguishing between a realizable
and an unrealizable sample. The core problem is that even if $S$ is
not realizable, the algorithm might still return a good
hypothesis. The crux of our new technique is the observation that if
$S$ is {\em randomly generated} unrealizable sample then even improper algorithm cannot return a
hypothesis with a small empirical error. The point is that the
returned hypothesis is determined solely by the examples that $\ca$
sees and its random bits. Therefore, if $\ca$ is an efficient
algorithm, the number of hypotheses it might return cannot be too
large. Hence, if $S$ is ``random enough", it likely to be far from all
these hypotheses, in which case the hypothesis returned by $\ca$ would
have a large error on $S$.

We now formalize this idea.
Let $\cd=\{\cd^{m(n)}_n\}_{n}$ be a polynomial ensemble of distributions, such that $\cd^{m(n)}_n$ is a distribution on $\cz_n^{m(n)}$. Think of $\cd^{m(n)}_n$ as a distribution that generates samples that are far from being realizable by $\ch$.
We say that it is hard to distinguish between a $\cd$-random sample and a realizable sample if there is no efficient randomized algorithm $\ca$ with the following properties:
\begin{itemize}
\item For every realizable sample $S\in \cz^{m(n)}_n$,
\[\Pr_{\text{internal coins of }\ca}\left(\ca(S)=``realizable"\right)\ge \frac{3}{4}~.\]
\item If $S\sim \cd_n^{m(n)}$, then with probability $1-o_n(1)$ over the choice of $S$, it holds that 
\[\Pr_{\text{internal coins of }\ca}\left(\ca(S)=``unrelizable"\right)\ge \frac{3}{4}~.\]
\end{itemize}
For functions $p,\epsilon:\mathbb N\to(0,\infty)$, we say that $\cd$ is {\em $(p(n),\epsilon(n))$-scattered} if, for large enough $n$, it holds that for every function $f:\cx_n\to\{0,1\}$,  
\[
\Pr_{S\sim\cd^{m(n)}_n}\left(\Err_{S}(f)\le \epsilon(n)\right)\le 2^{-p(n)}~.
\]
\begin{example}
Let $\cd^{m(n)}_n$ be the distribution over $\cz_n^{m(n)}$ defined by taking $m(n)$ independent uniformly chosen examples from $\cx_n\times\{0,1\}$. 
For $f:\cx_n\to\{0,1\}$, $\Pr_{S\sim\cd^{m(n)}_n}\left(\Err_{S}(f)\le \frac{1}{4}\right)$ is the probability of getting at most $\frac{m(n)}{4}$ heads in $m(n)$ independent tosses of a fair coin. By Hoeffding's bound, this probability is $\le 2^{-\frac{1}{8} m(n)}$. Therefore, $\cd=\{\cd^{m(n)}_n\}_{n}$ is $\left(\frac{1}{8} m(n),1/4\right)$-scattered.
\end{example}
\begin{theorem}\label{thm:basic_realizable}
Every hypothesis class that satisfies the following condition is not efficiently learnable. There exists $\beta>0$ such that for every $c>0$ there is an $(n^c,\beta)$-scattered ensemble $\cd$ for which it is hard to distinguish between a $\cd$-random sample and a realizable sample.
\end{theorem}
\begin{remark}
The theorem and the proof below work verbatim if we replace $\beta$ by $\beta(n)$, provided that $\beta(n)> n^{-a}$ for some $a>0$.
\end{remark}
\proof
Let $\ch$ be the hypothesis class in question and suppose toward a contradiction that algorithm $\cl$ learns $\ch$ efficiently. Let $M\left(n,1/\epsilon,1/\delta\right)$ be the maximal number of random bits used by $\cl$ when run on the input $n,\epsilon,\delta$. This includes both the bits describing the examples produced by the oracle and ``standard" random bits. Since $\cl$ is efficient, $M\left(n,1/\epsilon,1/\delta\right)< \poly(n,1/\epsilon, 1/\delta)$. Define
\[
q(n)=M\left(n,1/\beta,4\right)+n~.
\]
By assumption, there is a $(q(n),\beta)$-scattered ensemble $\cd$
for which it is hard to distinguish a $\cd$-random sample from a
realizable sample. Consider the algorithm $\ca$ defined below. On input $S\in\cz_n^{m(n)}$,
\begin{enumerate}
\item Run $\cl$ with parameters $n,\beta$ and $\frac{1}{4}$, such that the examples' oracle generates examples by choosing a random example from $S$.
\item Let $h$ be the hypothesis that $\cl$ returns. If $\Err_S(h)\le \beta$, output $\text{``realizable"}$. Otherwise, output $\text{``unrealizable"}$.
\end{enumerate}
Next, we derive a contradiction by showing that $\ca$ distinguishes a realizable sample from a $\cd$-random sample. Indeed, if the input $S$ is realizable, then $\cl$ is guaranteed to return, with probability $\ge 1-\frac{1}{4}$, a hypothesis $h:\cx_n\to\{0,1\}$ with $\Err_{S}(h)\le \beta$. Therefore, w.p. $\ge \frac{3}{4}$ $\ca$ will output ``realizable".

What if the input sample $S$ is drawn from $\cd^{m(n)}_n$? Let
$\cg\subset \{0,1\}^{\cx_n}$ be the collection of functions that $\cl$
might return when run with parameters $n,\epsilon(n)$ and
$\frac{1}{4}$. We note that $|\cg |\le 2^{q(n)-n}$, since  each
hypothesis in $\cg$ can be described by $q(n)-n$ bits. Namely, the
random bits that $\cl$ uses and the description of the examples
sampled by the oracle. Now, since $\cd$ is
$(q(n),\beta)$-scattered, the probability that $\Err_{S}(h)\le
\beta$ for some $h\in \cg$ is at most $|\cg|2^{-q(n)}\le
2^{-n}$. It follows that the probability that $\ca$ responds
``realizable" is $\le 2^{-n}$. This leads to the desired contradiction
and concludes our proof.
\proofbox

Next, we discuss analogue theorem to theorem \ref{thm:basic_realizable} for (approximate) agnostic learning. Let $\cd$ be a polynomial ensemble and $\epsilon:\mathbb N\to (0,1)$.
We say that it is hard to distinguish between a $\cd$-random sample and an $\epsilon$-almost realizable sample if there is no efficient randomized algorithm $\ca$ with the following properties:
\begin{itemize}
\item For every sample $S\in \cz^{m(n)}_n$ that is $\epsilon(n)$-almost realizable,
\[\Pr_{\text{internal coins of }\ca}\left(\ca(S)=``almost \;realizable"\right)\ge 3/4~.\]
\item If $S\sim \cd_n^{m(n)}$, then with probability $1-o_n(1)$ over the choice of $S$, it holds that 
\[\Pr_{\text{internal coins of }\ca}\left(\ca(S)=``unrelizable"\right)\ge \frac{3}{4}~.\]
\end{itemize}

\begin{theorem}\label{thm:basic_agnostic}
Let $\alpha \ge 1$. Every hypothesis class that satisfies the following condition is not efficiently agnostically learnable with an approximation ratio of $\alpha$.
For some $\beta$ and every $c>0$, there is a $(n^c,\alpha \beta+1/n)$-scattered ensemble $\cd$ such that it is hard to distinguish between a $\cd$-random sample and a $\beta$-almost realizable sample.
\end{theorem}
\begin{remark}
As in theorem \ref{thm:basic_realizable}, the theorem and the proof
below work verbatim if we replace $\alpha$ by $\alpha(n)$ and $\beta$ by $\beta(n)$, provided that $\beta(n)> n^{-a}$ for some $a>0$.
\end{remark}
\proof
Let $\ch$ be the hypothesis class in question and suppose toward a contradiction that $\cl$ efficiently agnostically learns $\ch$ with approximation ratio of $\alpha$. Let $M\left(n,1/\epsilon,1/\delta\right)$ be the maximal number of random bits used by $\cl$ when it runs on the input $n,\epsilon,\delta$. This includes both the bits describing the examples produced by the oracle and the ``standard" random bits. Since $\cl$ is efficient, $M\left(n,1/\epsilon,1/\delta\right)< \poly(n,1/\epsilon, 1/\delta)$. Define,
\[
q(n)=M\left(n,n,4\right)+n~.
\]
By the assumptions of the theorem, there is a $(q(n),\alpha\beta+1/n)$-scattered ensemble $\cd$ such that it is hard to distinguish between a $\cd$-random sample and a $\beta$-almost realizable sample. Consider the following efficient algorithm to distinguish between a $\cd$-random sample and a $\beta$-almost realizable sample. On input $S\in\cz_n^{m(n)}$,
\begin{enumerate}
\item Run $\cl$ with parameters $n,1/n$ and $\frac{1}{4}$, such that the examples are sampled uniformly from $S$.
\item Let $h$ be the hypothesis returned by the algorithm $\cl$. If $\Err_S(h)\le \alpha\beta+1/n$, return $\text{``almost realizable"}$. Otherwise, return  $\text{``unrealizable"}$.
\end{enumerate}
Next, we derive a contradiction by showing that this algorithm, which we denote by $\ca$, distinguishes between a realizable sample and a $\cd$-random sample. Indeed, if the input $S$ is $\beta$-almost realizable, then $\cl$ is guaranteed to return, with probability $\ge 1-\frac{1}{4}$, a hypothesis $h:\cx_n\to\{0,1\}$ with $\Err_{S}(h)\le \alpha\beta+1/n$. Therefore, the algorithm $\ca$ will return, w.p. $\ge \frac{3}{4}$, ``almost realizable".

Suppose now that the input sample $S$ is drawn according to $\cd_n$. Let $\cg\subset \{0,1\}^{\cx_n}$ be the collection of functions that the learning algorithm $\cl$ might return when it runs with the parameters $n,1/n$ and $\frac{1}{4}$. Note that each hypothesis in $\cg$ can be described by $q(n)-n$ bits, namely, the random bits used by $\cl$ and the description of the examples sampled by the oracle. Therefore, $|\cg |\le 2^{q(n)-n}$. Now, since $\cd$ is $(q(n),\alpha\beta+1/n)$-scattered, the probability that some function in $h\in \cg$ will have $\Err_{S}(h)\le \alpha\beta+1/n$ is at most $|\cg|2^{-q(n)}\le 2^{-n}$. It follows that the probability that the algorithm $\ca$ will return ``almost realizable" is $\le 2^{-n}$.
\proofbox

\section{The strong random CSP assumption}
In this section we put forward and discuss a new assumption that we call ``the strong random CSP assumption" or $\srcsp$ for short. It generalizes Feige's assumption \citep{Feige02}, as well as the assumption of Barak, Kindler and Steurer \citep{BarakKiSt13}. This new assumption, together with the methodology described in section \ref{sec:methodology}, are used to establish lower bounds for improper learning. Admittedly, our assumption is strong, and an obvious quest, discussed in the end of this section is to find ways to derive similar conclusions from weaker assumptions.

The $\srcsp$ assumption claims that for certain predicates $P:\{\pm 1\}^K\to \{0,1\}, d>0$ and $\alpha>0$, the decision problem $\csp^{\alpha,\rand}_{n^d}(P)$ is intractable. We first consider the case $\alpha=1$.
To reach a plausible assumption, let us first discuss Feige's assumption, and the existing evidence for it. Denote by $\sat_3:\{\pm 1\}^3\to \{0,1\}$ the $3$-$\sat$ predicate $\sat_3(x_1,x_2,x_3)=x_1\vee x_2\vee x_3$. 
\begin{assumption}[Feige]\label{hyp:feige}
For every sufficiently large constant $C>0$, $\csp^{1,\rand}_{C\cdot n}(\sat_3)$ is intractable.
\end{assumption}
Let us briefly summarize the evidence for this assumption.
\begin{itemize}
\item {\bf Hardness of approximation.} Feige's conjecture can be viewed as a
strengthening of Hastad's celebrated result \citep{haastad2001some} that $\sat_3$ is approximation resistant on satisfiable instances. Hastad's result implies that under $\mathbf{P}\ne \mathbf{NP}$, it is hard to distinguish
satisfiable instances to $\csp(\sat_3)$ from instances with value $\le \frac{7}{8}+\epsilon$. The collection of instances with value $\le \frac{7}{8}+\epsilon$ includes most random instances with $C\cdot n$ clauses for sufficiently
large $C$. Feige's conjecture says that the problem remains intractable even when restricted to these random instances.

We note that approximation resistance on satisfiable instances is a necessary condition for the validity of Feige's assumption. Indeed, for large enough $C>0$, with probability $1-o_n(1)$, the value of a random instance to $\csp(\sat_3)$ is $\le \frac{7}{8}+\epsilon$. Therefore, tractability of $\csp^{1,\frac{7}{8}+\epsilon}(\sat_3)$ would lead to tractability of $\csp^{1,\rand}_{C\cdot n}(\sat_3)$.

\item {\bf Performance of known algorithms.} The problem of refuting random $3$-$\sat$ formulas has been extensively studied and a many algorithms were studied. The best known algorithms \citep{feige2004easily} can refute random instances with $\Omega\left(n^{1.5}\right)$ random constraints. Moreover resolution lower bounds \citep{BenWi99} show that many algorithms run for exponential time when applied to random instances with $O\left(n^{1.5-\epsilon}\right)$ constraints.
\end{itemize}
We aim to generalize Feige's assumption in two aspects -- (i) To predicates other than $\sat_3$, and (ii) To problems with super-linearly many constraints. Consider the problem $\csp^{1,\rand}_{m(n)}(P)$ for some predicate $P:\{\pm 1\}^K\to\{0,1\}$.
As above, the intractability of $\csp^{1,\rand}_{m(n)}(P)$ strengthens the claim that $P$ is approximation resistant on satisfiable instances. Also, for $\csp^{1,\rand}_{m(n)}(P)$ to be hard, it is necessary that $P$ is approximation resistant on satisfiable instances.
In fact, as explained in section \ref{sec:preliminaries_res}, 
if $P':\{\pm 1\}^K\to\{0,1\}$ is implied by $P$, then the problem $\csp^{1,\rand}_{m(n)}(P)$ can be easily reduced to $\csp^{1,\rand}_{m(n)}(P')$. Therefore, to preserve the argument of the first evidence of Feige's conjecture, it is natural to require that $P$ is {\em heredity} approximation resistant on satisfiable instances.

Next, we discuss what existing algorithms can do. The best known algorithms for the predicate $\sat_K(x_1,\ldots,x_K)=\vee_{i=1}^Kx_i$ can only refute random instances with $\Omega\left(n^{\lfloor\frac{K}{2}\rfloor}\right)$ constraints \citep{coja2010efficient}.
This gives some evidence that it becomes harder to refute random instances of $\csp(P)$  as the number of variables grows. Namely, that many random constraints are needed to efficiently refute random instances. Of course, some care is needed with counting the ``actual" number of variables. Clearly, only {\sl certain} predicates have been studied so far. Therefore, to reach a plausible assumption, we consider the {\em resolution refutation complexity} of random instances to $\csp(P)$. And consequently, also the performance of a large class of algorithms, including Davis-Putnam style (DPLL) algorithms. 

Davis-Putnam algorithms have been subject to an extensive study, both theoretical and empirical. Due to the central place that they occupy, much work has been done since the late 80's, to prove lower bounds on their performance in refuting random $K$-$\sat$ formulas. These works relied on the fact that these algorithms implicitly produce a resolution refutation during their execution. Therefore, to derive a lower bound on the run time of these algorithms, exponential lower bounds were established on the resolution complexity of random instances to $\csp(\sat_K)$. These lower bounds provide support to the belief that it is hard to refute not-too-dense random $K$-$\sat$ instances.

We define the {\em $0$-variability,} $\var_0(P)$, of a predicate $P$ as the smallest cardinality of a set of $P$'s variables such that there is an assignment to these variables for which $P(x)=0$, regardless of the values assigned to the other variables. By a simple probabilistic argument, a random $\csp(P)$ instance with $\Omega\left(n^r\right)$ constraints, where $r=\var_0(P)$ is almost surely unsatisfiable with a resolution proof of constant size. Namely, w.p. $1-o_n(1)$, there are $2^r$ constraints that are inconsistent, since some set of $r$ variables appears in all $2^r$ possible ways in the different clauses.
On the other hand, we show in section \ref{sec:res} that a random $\csp(P)$ problem with  $O\left(n^{c\cdot r}\right)$ constraints has w.h.p. exponential resolution complexity. Here $c>0$ is an absolute constant. Namely,
\begin{theorem}\label{thm:res_main}
There is a constant $C>0$ such that for every $d>0$ and every predicate $P$ with $\var_0(P)\ge C\cdot d$, the following holds. With probability $1-o_n(1)$, a random instance of $\csp(P)$ with $n$ variables and $n^d$ constraints has resolution refutation length $\ge 2^{\Omega\left(\sqrt{n}\right)}$.
\end{theorem}

To summarize, we conclude that the parameter $\var_0(P)$ controls the resolution complexity of random instances to $\csp(P)$. 
In light of the above discussion, we put forward the following assumption.

\begin{assumption}[SRCSP -- part 1]\label{hyp:sat}
There is a function $f:\mathbb N\to\mathbb N$ such that the following holds. Let $P$ be a predicate that is heredity approximation resistant on satisfiable instances with
$\var_0(P)\ge f(d)$.
Then, it is hard to distinguish between satisfiable instances of $\csp(P)$ and random instances with $n^d$ constraints.
\end{assumption}
Next, we motivate a variant of the above assumption, that accommodates also predicates that are not heredity approximation resistant. A celebrated result of Raghavendra \citep{raghavendra2008optimal} shows that under the unique games conjecture \citep{khot2002power}, a certain $\mathrm{SDP}$-relaxation-based algorithm is (worst case) optimal for $\csp(P)$, for every predicate $P$. Barak et al. \cite{BarakKiSt13} conjectured that this algorithm is optimal even on random instances. They considered the performance of this algorithm on random instances and purposed the following assumption, which they called the ``random $\csp$ hypothesis". Define $\Uval(P)=\max_{\cd}\E_{x\sim\cd}P(x)$, where the maximum is taken over all pairwise uniform distributions\footnote{A distribution is {\em pairwise uniform} if, for every pair of coordinates, the distribution induced on these coordinates is uniform.} on $\{\pm 1\}^K$.

\begin{assumption}[RSCP]\label{hyp:barak_kindler_steurer}
For every $\epsilon>0$ and sufficiently large $C>0$, it is hard to distinguish instances with value $\ge \Uval(P)-\epsilon$ from random instances with $C\cdot n$ constraints.
\end{assumption}
Here we generalize the $\rcsp$ assumption to random instances with much more than $C\cdot n$ constraints. As in assumption \ref{hyp:sat}, the $0$-variability of $P$ serves to quantify the number of random constraints needed to efficiently show that a random instance has value $< \Uval(P)-\epsilon$.

\begin{assumption}[SRSCP - part 2]\label{hyp:not_sat}
There is a function $f:\mathbb N\to\mathbb N$ such that for every predicate $P$ with $\var_0(P)\ge f(d)$ and for every $\epsilon>0$, it is hard to distinguish between instances with value $\ge \Uval(P)-\epsilon$ and random instances with $n^d$ constraints.
\end{assumption}
Finally, we define the notion of a $\srcsp$-hard problem.
\begin{terminology}
A computational problem is $\srcsp$-hard if its tractability contradicts assumption \ref{hyp:sat} or \ref{hyp:not_sat}.
\end{terminology}
\subsection{Toward weaker assumptions}\label{sec:weaker}
The $\srcsp$ assumption is strong. It is highly desirable to arrive at similar conclusions from substantially weaker assumptions. A natural possibility that suggests itself is the $\srcsp$ assumption, restricted to $\sat$:
\begin{assumption}\label{hyp:only_sat}
There is a function $f:\mathbb N\to\mathbb N$ such that for every $K\ge f(d)$, it is hard to distinguish satisfiable instances of $\csp(\sat_K)$ from random instances with $n^d$ constraints.
\end{assumption}
We are quite optimistic regarding the success of this direction: The lower bounds we prove here use the $\srcsp$-assumption only for certain predicates, and do not need the full power of the assumption. Moreover, for the hypothesis classes of $\dnf$'s, intersection of halfspaces, and finite automata, these predicates are somewhat arbitrary. In \citep{Feige02}, it is shown that for predicates of arity $3$, assumption \ref{hyp:not_sat} is implied by the same assumption restricted to the $\sat$ predicate.
This gives a hope to prove, based on assumption \ref{hyp:only_sat}, that the $\srcsp$-assumption is true for predicates that are adequate to our needs.

\section{Summary of results}
\subsection{Learning $\dnf$'s}
A {\em $\dnf$ clause} is a conjunction of literals. A {\em $\dnf$ formula} is a disjunction of $\dnf$ clauses. Each $\dnf$ formula over $n$ variables naturally induces a function on $\{\pm 1\}^n$.  
We define the size of a $\dnf$ clause as the number of its literals and the size of a $\mathrm{DNF}$ formula as the sum of the sizes of its clauses.

As $\dnf$ formulas are very natural form of predictors, learning
hypothesis classes consisting of $\dnf$'s formulas of polynomial size
has been a major effort in computational learning theory. Already in
Valiant's paper \citep{Valiant84}, it is shown that for every constant
$q$, the hypothesis class of all $\dnf$-formulas with $\le q$ clauses
is efficiently learnable. The running time of the algorithm is,
however, exponential in $q$. We also note that Valiant's algorithm is
improper. For general polynomial-size $\dnf$'s, the best known result
\citep{klivans2001learning} shows learnability in time $\frac{1}{\epsilon}\cdot
2^{\tilde{O}\left(n^{\frac{1}{3}}\right)}$. Better running times
(quasi-polynomial) are known under distributional assumptions
\citep{LinialMaNi89,mansour1995nlog}.

As for lower bounds, {\em properly} learning $\dnf$'s is known to be
hard \citep{PittVa88}. However, proving hardness of improper learning
of polynomial $\dnf$'s has remained a major open question in
computational learning theory. Noting that $\dnf$ clauses coincide
with depth $2$ circuits, a natural generalization of $\dnf$'s is
circuits of small depth.  For such classes, certain lower bounds can
be obtained using the cryptographic technique. Kharitonov
\cite{Kharitonov93} has shown that a certain subexponential lower
bound on factoring Blum integers implies hardness of learning circuits
of depth $d$, for some unspecified constant $d$. Under more standard assumptions
(that the $\rsa$ cryptosystem is secure), best lower bounds
\citep{KearnsVa89} only rule out learning of circuits of depth
$\log(n)$.

For a function $q:\mathbb N\to\mathbb N$, denote by
$\mathrm{DNF}_{q(n)}$ the hypothesis class of functions over $\{\pm
1\}^n$ that can be realized by $\mathrm{DNF}$ formulas of size at most
$q(n)$.  Also, let $\mathrm{DNF}^{q(n)}$ be the hypothesis class of
functions over $\{\pm 1\}^n$ that that can be realized by
$\mathrm{DNF}$ formulas with at most $q(n)$ clauses. Since each clause
is of size at most $n$, $\mathrm{DNF}^{q(n)}\subset
\mathrm{DNF}_{nq(n)}$.

As mentioned, for a constant $q$, the class $\mathrm{DNF}^{q}$ is efficiently learnable. We show that for every super constant $q(n)$, it is $\srcsp$-hard to learn $\mathrm{DNF}^{q(n)}$:
\begin{theorem}\label{thm:dnf_few_clauses}
If $\lim_{n\to\infty}q(n)=\infty$ then learning $\mathrm{DNF}^{q(n)}$ is $\srcsp$-hard.
\end{theorem}
Since $\mathrm{DNF}^{q(n)}\subset \mathrm{DNF}_{nq(n)}$, we immediately conclude that learning $\dnf$'s of size, say, $\le n\log(n)$, is $\srcsp$-hard. By a simple scaling argument, we obtain an even stronger result:
\begin{corollary}\label{cor:dnf_small}
For every $\epsilon>0$, it is $\srcsp$-hard to learn $\dnf_{n^\epsilon}$.
\end{corollary}
\begin{remark}\label{rem:boosting}
  Following the Boosting argument of Schapire \cite{Schapire89},
  hardness of improper learning of a class $\ch$ immediately implies
  that for every $\epsilon > 0$, there is no efficient algorithm that
  when running on a distribution that is realized by $\ch$, guaranteed
  to output a hypothesis with error $\le
  \frac{1}{2}-\epsilon$. Therefore, hardness results of improper
  learning are very strong, in the sense that they imply that the
  algorithm that just makes a random guess for each example, is
  essentially optimal.
\end{remark}

\subsection{Agnostically learning halfspaces}
Let $\half$ be the hypothesis class of halfspaces over
$\{-1,1\}^n$. Namely, for every $w\in\mathbb R^n$ we define $h_w:\{\pm
1\}^n\to\{0,1\}$ by $h_w(x)=\sign\left(\inner{w,x}\right)$, and let
\[
\half=\left\{h_w\mid w\in\mathbb R^n\right\}~.
\]
We note that usually halfspaces are defined over $\mathbb R^n$, but
since we are interested in lower bounds, looking on this more
restricted class just make the lower bounds stronger.

The problem of learning halfspaces is as old as the field of machine
learning, starting with the perceptron algorithm \citep{Rosenblatt58},
through the modern $\mathrm{SVM}$ \citep{Vapnik98}. 
As opposed to learning $\dnf$'s, learning halfspaces in the realizable
case is tractable. However, in the agnostic PAC model, 
the best currently known algorithm for learning halfspaces runs in
time exponential in $n$ and the best known approximation ratio of
polynomial time algorithms is
$O\left(\frac{n}{\log(n)}\right)$. Better running times (usually of
the form $n^{\poly\left(\frac{1}{\epsilon}\right)}$) are known under
distributional assumptions (e.g. \cite{kalai2008agnostically}).

The problem of \emph{proper} agnostic learning of halfspaces was
shown to be hard to approximate within a factor of
$2^{\log^{1-\epsilon}(n)}$ \citep{arora1993hardness}.  Using the
cryptographic technique, improper learning of halfspaces is known to
be hard, under a certain cryptographic assumption regarding the
shortest vector problem (\cite{FeldmanGoKhPo06}, based on
\cite{KlivansSh06}).  No hardness results are known for \emph{approximately
and improperly} learning halfspaces. Here, we show that:
\begin{theorem}\label{thm:halfspaces}
  For every constant $\alpha\ge 1$, it is $\srcsp$-hard to
  approximately agnostically learn $\half$ with an approximation ratio
  of $\alpha$.
\end{theorem}

\subsection{Learning intersection of halfspaces}
For a function $q:\mathbb N\to\mathbb N$, we let $\inter_{q(n)}$ be the hypothesis class of intersection of $\le q(n)$ halfspaces. That is, 
$\inter_{q(n)}$ consists of all functions $f:\{\pm 1\}^n\to\{0,1\}$ 
for which there exist $w_1,\ldots w_k\in\mathbb R^{n}$ such that $f(x)=1$ if and only if $\forall i,\inner{w_i,x}>0$. 

Learning intersection of halfspaces has been a major challenge in
machine learning. Beside being a natural generalization of learning
halfspaces, its importance stems from {\em neural networks}
\citep{bishop1995neural}. Learning neural networks was popular in the
80's, and enjoy a certain comeback nowadays. A neural network is
composed of layers, each of which is composed of nodes. The first
layer consists of $n$ nodes, containing the input values.  The nodes
in the rest of the layers calculates a value according to a halfspace
(or a ``soft'' halfspace obtained by replacing the sign function with
a sigmoidal function) applied on the values of the nodes in the
previous layer. The final layer consists of a single node, which is
the output of the whole network.

Neural networks naturally induce several hypothesis classes (according
to the structure of the network). The class of intersection of
halfspaces is related to those classes, as it can be realized by very
simple neural networks: the class $\inter_{q(n)}$ can be realized by
neural networks with only an input layer, a single hidden layer, and
output layer, so that there are $q(n)$ nodes in the second layer.
Therefore, lower bounds on improperly learning intersection of halfspaces implies lower
bounds on improper learning of neural networks.

Exact algorithms for learning $\inter_{q(n)}$ run in time exponential
in $n$. Better running times (usually of the form
$n^{\poly\left(\frac{1}{\epsilon}\right)}$) are known under
distributional assumptions (e.g. \cite{klivans2002learning}).  It is
known that properly learning intersection of even 2 halfspaces is hard
\citep{khot2011hardness}. For improper learning, Klivans and Sherstov
\cite{KlivansSh06} have shown that learning an intersection of
polynomially many half spaces is hard, under a certain cryptographic
assumption regarding the shortest vector problem.  Noting that every
$\dnf$ formula with $q(n)$ clauses is in fact the complement of an
intersection of $q(n)$ halfspaces\footnote{In the definition of $\inter$, we considered halfspaces with no threshold, while halfspaces corresponding to $\dnf$s do have a threshold. This can be standardly handled by padding the examples with a single coordinate of value $1$.}, we conclude from theorem
\ref{thm:dnf_few_clauses} that intersection of every super constant
number of halfsapces is hard.
\begin{theorem}\label{thm:intersection}
If $\lim_{n\to\infty}q(n)=\infty$ then learning $\inter_{q(n)}$ is $\srcsp$-hard.
\end{theorem}
In section \ref{sec:4half} we also describe a route that might lead to the result that learning $\inter_{4}$ is $\srcsp$-hard.

\subsection{Additional results}
In addition to the results mentioned above, we show that learning the class of finite automata of polynomial size is $\srcsp$-hard. Hardness of this class can also be derived using the cryptographic technique, based on the assumption that the $\rsa$ cryptosystem is secure \citep{KearnsVa89}.
Finally, we show that agnostically learning parity with any constant approximation ratio is $\srcsp$-hard. Parity is not a very interesting class from the point of view of practical machine learning. However, learning this class is related to several other problems in complexity \citep{blum2003noise}. We note that hardness of agnostically learning parity, even in a more relaxed model than the agnostic PAC model (called the random classification noise model), is a well accepted hardness assumption.

In section \ref{sec:res} we prove lower bounds on the size of a resolution refutation for random $\csp$ instances. In section \ref{sec:base_on_np} we show that unless the polynomial hierarchy collapses, there is no ``standard reduction" from an $\mathbf{NP}$-hard problem (or a $\mathbf{CoNP}$-hard problem) to random $\csp$ problems.

\subsection{On the proofs}
Below we outline the proof for $\dnf$s. The proof for halfspaces and
parities is similar. For every $c>0$, we start with a predicate
$P:\{\pm 1\}^K\to\{0,1\}$, for which the problem
$\csp^{1,\rand}_{n^c}(P)$ is hard according to the
$\srcsp$-assumption, and reduce it to the problem of distinguishing
between a $(\Omega(n^c),\frac{1}{5})$-scattered sample and a
realizable sample. Since $c$ is arbitrary, the theorem follows from
theorem \ref{thm:basic_realizable}.

The reduction is performed as follows. Consider the problem
$\csp(P)$. Each assignment naturally defines a function from the
collection of $P$-constraints to $\{0,1\}$. Hence, if we think about
the constraints as instances and about the assignments as hypotheses,
the problem $\csp(P)$ turns into some kind of a learning
problem. However, in this interpretation, all the instances we see
have positive labels (since we seek an assignment that satisfies as
many instances as possible). Therefore, the problem
$\csp^{1,\rand}_{n^c}(P)$ results in ``samples" which are not
scattered at all.

To overcome this, we show that the analogous problem to
$\csp^{1,\rand}_{n^c}(P)$, where $(\neg P)$-constraints are also
allowed, is hard as well (using the assumption on the hardness of
$\csp^{1,\rand}_{n^c}(P)$). The hardness of the modified problem can
be shown by relying on the special predicate we
work with. This predicate was defined in the recent work of Huang
\citep{huang2013approximation}, and it has the property of being
heredity approximation resistant, even though $|P^{-1}(1)|\le
2^{O\left(K^{1/3}\right)}$.

At this point, we have an (artificial) hypothesis class which is
$\srcsp$-hard to learn by theorem \ref{thm:basic_realizable}. In the
next and final step, we show that this class can be efficiently
realized by $\dnf$s with $\omega(1)$ clauses. The reduction uses the
fact that every boolean function can be expressed by a $\dnf$ formula
(of possibly exponential size). Therefore, $P$ can be expressed by a
$\dnf$ formula with $2^K$ clauses. Based on this, we show that each
hypothesis in our artificial class can be realized by a $\dnf$ formula
with $2^K$ clauses, which establishes the proof.

The results about learning automata and learning intersection of $\omega(1)$ halfspaces follow from the result about $\dnf$s: We show that these classes can efficiently realize the class of $\dnf$s with $\omega(1)$ clauses. In section \ref{sec:4half} we suggest a route that might lead to the result that learning intersection of $4$ halfspaces is $\srcsp$-hard: We show that assuming the unique games conjecture, a certain family of predicates are heredity approximation resistant. We show also that for these predicates, the problem $\csp^{1,\alpha}(P)$ is $\mathbf{NP}$-hard for some $1>\alpha>0$. This leads to the conjecture that these predicates are in fact heredity approximation resistant. Conditioning on the correctness of this conjecture, we show that it is $\srcsp$-hard to learn intersection of $4$-halfspaces. This is done using the strategy described for $\dnf$s.

The proof of the resolution lower bounds (section \ref{sec:res}) relies on the strategy and the ideas introduced in \citep{haken1985intractability} and farther developed in \citep{BeamePi96,BeameKaPiSa98,BenWi99}.
The proof that it is unlikely that the correctness of the $\srcsp$-assumption can be based on $\mathbf{NP}$-hardness (section \ref{sec:base_on_np}) uses the idea introduced in \citep{ApplebaumBaXi08}: we show that if an $\mathbf{NP}$-hard problem (standardly) reduces to $\csp^{\alpha,\rand}_{m(n)}(P)$, then the problem has a statistical zero knowledge proof. It follows that $\mathbf{NP}\subset \mathbf{SZKP}$, which collapses the polynomial hierarchy.

\section{Future work}\label{sec:future}
We elaborate below on some of the numerous open problems and research directions that the present paper suggests.

\subsection{Weaker assumptions?}
First and foremost, it is very
desirable to draw similar conclusions from assumption substantially weaker than $\srcsp$
(see section~\ref{sec:weaker}). Even more ambitiously, is it possible to reduce some $\mathbf{NP}$-hard
problem to some of the problems that are deemed hard by the
$\srcsp$ assumption? In section \ref{sec:base_on_np}, we show that a pedestrian application of this approach is doomed to fail (unless the
polynomial hierarchy collapses). This provides, perhaps, a
moral justification for an ``assumption based" study of average case
complexity.

\subsection{The $\srcsp$-assumption}
We believe that the results presented here, together with \citep{Feige02,Alekhnovich03,daniely2013more,berthet2013computational} and \citep{BarakKiSt13}, make a compelling case that it is of fundamental importance for complexity theory to understand the hardness of random $\csp$ problems. In this context, the $\srcsp$ assumption is an interesting conjecture. There are, of course, many ways to try to refute it. On the other hand, current techniques in complexity theory seem too weak to prove it, or even to derive it from standard complexity assumptions. Yet, there are ways to provide more circumstantial evidence in favor of this assumption:
\begin{itemize}
\item As discussed in the previous section, one can try to derive it, even partially, from weaker assumptions. 
\item Analyse the performance of existing algorithms. In section
  \ref{sec:res} it is shown that no Davis-Putnam algorithm can
  refute the $\srcsp$ assumption. Also, Barak et al \citep{BarakKiSt13} show that the basic $\mathbf{SDP}$ algorithm \citep{raghavendra2008optimal} cannot refute assumption \ref{hyp:not_sat}, and also \ref{hyp:sat} for certain predicates (those that contain a pairwise uniform distribution).
Such results regarding additional classes of algorithms will
lend more support to the assumption's correctness.
\item Show lower bounds on the proof complexity of random $\csp$ instances in refutation systems stronger than resolution.
\end{itemize}
For a further discussion, see
\cite{BarakKiSt13}. Interest in the
$\srcsp$ assumption calls for a better understanding of heredity approximation
resistance. For recent work in this direction, see \citep{huang2012approximation,huang2013approximation}.

\subsection{More applications}
We believe that the method presented here and the $\srcsp$-assumption
can yield additional results in learning and approximation. Here
are several basic questions in learning theory that we are unable to resolve even
under the $\srcsp$-assumption.
\begin{enumerate}
\item Decision trees are very natural hypothesis class, that is not known to be efficiently learnable. Is it $\srcsp$-hard to learn decision trees?
\item What is the real approximation ratio of learning halfspaces? We showed that it is $\srcsp$-hard to agnostically learn halfspaces with a constant approximation ratio. The best known algorithm only guarantees an approximation ratio of $\frac{n}{\log n}$. This is still a huge gap. See remark \ref{rem:halfsapces} for some speculations about this question.
\item Likewise for learning large margin halfspaces (see remark \ref{rem:large_margin}) and for parity.
\item Prove that it is $\srcsp$-hard to learn intersections of a constantly many halfspaces. This might be true even for $2$ halfspaces. In section \ref{sec:4half}, we suggest a route to prove that intersection of $4$ halfspaces is $\srcsp$-hard.
\end{enumerate}
Besides application to learning and approximation, it would be
fascinating to see applications of the $\srcsp$-assumption in other
fields of complexity. It will be a poetic justice if we could apply it to
cryptography. We refer the reader to \citep{BarakKiSt13} for a
discussion. Finding implications in fields beyond cryptography,
learning and approximation would be even more exciting.

\section{Proofs of the lower bounds}\label{sec:proofs}

Relying on our general methodology given in section
\ref{sec:methodology}, to show that a learning problem is
$\srcsp$-hard, we need to find a scattered ensamble, $\cd$, such that
it is $\srcsp$-hard to distinguish between a realizable sample and a
$\cd$-random sample. We will use the following simple criterion for an
ensamble to be scattered.
\begin{proposition}\label{prop:scat_crit}
  Let $\cd$ be some distribution on a set $\cx$.  For even $m$, let
  $X_1,\ldots,X_m$ be independent random variables drawn according to
  $\cd$. Consider the sample
  $S=\{(X_1,1),(X_2,0)\ldots,(X_{m-1},1),(X_{m},0)\}$. Then, for every
  $h:\cx\to\{0,1\}$,
\[
\Pr_{S}\left(\Err_S(h)\le \frac{1}{5}\right)\le 2^{-\frac{9}{100}m}
\]
\end{proposition}
\proof
For $1\le i\le \frac{m}{2}$ let $T_i=1[h(X_{2i-1})\ne 1]+1[h(X_{2i})\ne 0]$. Note that 
$\Err_{S}(h)=\frac{1}{m}\sum_{i=1}^{\frac{m}{2}} T_i$. Also, the $T_i$'s are independent random variables with mean $1$ and values between $0$ and $2$. Therefore, by Hoeffding's bound,
\[
\Pr_{S}\left(\Err_S(h)\le \frac{1}{5}\right)\le e^{-\frac{9}{100}m} \le 2^{-\frac{9}{100}m}~.
\]
\proofbox

\subsection{Learning DNFs}
In this section we prove theorem \ref{thm:dnf_few_clauses} and
corollary \ref{cor:dnf_small}. We will use the $\srcsp$ assumption
\ref{hyp:sat} with Huang's predicate
\citep{huang2013approximation}. Let $k\ge 1$ and denote $K=k+\binom
k3$. We index the first $k$ coordinates of vectors in $\{\pm 1\}^{K}$
by the numbers $1,2,\ldots,k$. The last $\binom{k}{3}$ coordinates are
indexed by $\binom{[k]}{3}$. Let $H_k:\{\pm 1\}^{K}\to \{0,1\}$ be the
predicate such that $H_k(x)=1$ if and only if there is a vector $y$
with hamming distance $\le k$ from $x$ such that, for every $A\in
\binom{[k]}{3}$, $y_A=\prod_{i\in A}y_i$. The basic properties of
$H_k$ are summarized in the following lemma due to \citep{huang2013approximation}.
\begin{lemma}[\citep{huang2013approximation}]\label{lem_Huang}
\
\begin{enumerate} 
\item $H_k$ is heredity approximation resistant on satisfiable instances. 
\item $|H_k^{-1}(1)|=\tilde{O}(K^{1/3})$.
\item The $0$-variability of $H_k$ is $\ge k$.
\item For every sufficiently large $k$, there exists $y^k\in\{\pm 1\}^{K}$ such that $H_k(x)=1\Rightarrow H_k(y^k\oplus x)=0$
\end{enumerate}
\end{lemma}
\proof
1. and 2. were proved in \cite{huang2013approximation}. 3. is very easy. We proceed to 4. Choose $y^k\in \{\pm 1\}^K$ uniformly at random. By 2., for every $x\in \{\pm 1\}^K$, $\Pr\left(H_k(y^k\oplus x)=1\right)=2^{-K+\tilde{O}(K^{1/3})}$. Taking a union over all vectors $x\in H_k^{-1}(1)$, we conclude that the probability that one of them satisfies $H_k(y^k\oplus x)=1$ is $2^{-K+\tilde{O}(K^{1/3})+\tilde{O}(K^{1/3})}=2^{-K+\tilde{O}(K^{1/3})}$. For large enough $k$, this is less than $1$. Therefore, there exists a $y^k$ as claimed.
\proofbox

\proof (of theorem \ref{thm:dnf_few_clauses})
Let $d>0$ by assumption \ref{hyp:sat} and lemma \ref{lem_Huang}, for
large enough $k$, it is $\srcsp$-hard to distinguish between
satisfiable instances to $\csp(H_k)$ and random instances with
$m=2n^d$ constraints. We will reduce this problem to the problem of
distinguishing between a realizable sample to $\dnf^{q(n)}$ and a
random sample drawn from a
$\left(\frac{9}{50}n^d,1/5\right)$-scattered ensamble $\cd$. Since $d$ is arbitrary, the theorem follows from theorem \ref{thm:basic_realizable}.

The reduction works as follows. Let $y^k$ be the vector from lemma \ref{lem_Huang}. Given an instance
\[
J=\{H_k(j_{1,1}x_{i_{1,1}},\ldots,j_{1,K}x_{i_{1,K}}),
\ldots,H_k(j_{m,1}x_{i_{m,1}},\ldots,j_{m,K}x_{i_{m,K}})\}
\]
to $\csp(H_k)$, we will produce a new instance $J'$ by changing the sign of the variables according to $y^k$ in every other constraint. Namely,
\begin{eqnarray*}
J'&=&\{H_k(j_{1,1}x_{i_{1,1}},\ldots,j_{1,K}x_{i_{1,K}}),
H_k(y^k_1j_{2,1}x_{i_{2,1}},\ldots,y^k_Kj_{2,K}x_{i_{2,K}}),
\ldots
\\
&&\ldots, H_k(j_{m-1,1}x_{i_{m-1,1}},\ldots,j_{m-1,K}x_{i_{m-1,K}}),
H_k(y^k_1j_{m,1}x_{i_{m,1}},\ldots,y^k_Kj_{m,K}x_{i_{m,K}})\}~.
\end{eqnarray*}
Note that if $J$ is random then so is $J'$. Also, if $J$ is satisfiable with a satisfying assignment $u$, then, by lemma \ref{lem_Huang}, $u$ satisfies in $J'$ exactly the constraints with odd indices. Next, we will produce a sample $S\in\left(\{\pm 1\}^{2Kn}\times \{0,1\}\right)^{m}$ from $J'$ as follows. We will index the coordinates of vectors in $\{\pm 1\}^{2Kn}$ by $[K]\times \{\pm 1\}\times[n]$. We define a mapping $\Psi$ from the collection of $H_k$-constraints to $\{\pm 1\}^{2Kn}$ as follows -- for each constraint $C=H_k(j_1x_{i_1},\ldots,j_Kx_{i_K})$ we define $\Psi(C)\in \{\pm 1\}^{2Kn}$ by the formula
\[
\left(\Psi(C)\right)_{l,b,i}=\begin{cases}
-1 & (b,i)=(-j_l,i_l)
\\
1 & \textrm{otherwise}
\end{cases}
\]
Finally, if $J'=\{C'_1,\ldots,C'_m\}$, we will produce the sample
\[
S=\left\{(\Psi(C'_1),1),(\Psi(C'_2),0),\ldots,(\Psi(C'_{m-1}),1),(\Psi(C'_m),0)\right\} ~.
\]
The theorem follows from the following claim:
\begin{claim}
\
\begin{enumerate}
\item If $J$ is a random instance then $S$ is $\left(\frac{9}{100}m,\frac{1}{5}\right)$-scattered.
\item If $J$ is a satisfiable instance then $S$ is realizable by a $\dnf$ formula with $\le 2^K$ clauses.
\end{enumerate}
\end{claim}
Proposition \ref{prop:scat_crit} implies part 1. We proceed to part 2. Like every boolean function on $K$ variables, $H_k$ is expressible by a $\dnf$ expression of $2^K$ clauses, each of which contains all the variables. Suppose then that
\[
H_k(x_1,\ldots,x_K)=\vee_{t=1}^{2^K}\wedge_{r=1}^Kb_{t,r}x_r~.
\] 
Let $u\in \{\pm 1\}^n$ be an assignment to $J$. Consider the following $\dnf$ formula over $\{\pm 1\}^{2Kn}$
\[
\phi_{u}(x)=\vee_{t=1}^{2^K}\wedge_{r=1}^K\wedge_{i=1}^n x_{r,(u_i b_{t,r}),i}~,
\] 
where, as mentioned before, we index coordinates of $x \in \{\pm
1\}^{2Kn}$ by triplets in $[K] \times \{\pm 1\} \times [n]$. 
We claim that for every $H_k$-constraint $C$, $\phi_u(\Psi(C))=C(u)$. This suffices, since if $u$ satisfies $J$ then $u$ satisfies exactly the constraints with odd indices in $J'$. Therefore, by the definition of $S$ and the fact that $\forall C, \phi_u(\Psi(C))=C(u)$, $\phi_{u}$ realizes $S$.

Indeed, let $C(x)=H_k(j_1x_{i_1},\ldots,j_Kx_{i_K})$ be a
$H_k$-constraint. 
We have 
\begin{align*}
\phi_u(\Psi(C))=1 ~~&\iff \exists t\in[2^K]\,\forall r\in[K], i\in [n],\;
(\Psi(C))_{r,(u_ib_{t,r}),i}=1 \\
&\iff \exists t\in[2^K]\,\forall r\in[K], i\in [n]\;
(u_ib_{t,r},i)\ne(-j_r,i_r)  \\
&\iff \exists t\in[2^K]\,\forall r\in[K],\; u_{i_r}b_{t,r}\ne -j_r \\
&\iff \exists t\in[2^K]\,\forall r\in[K],\; b_{t,r}= j_ru_{i_r}  \\
&\iff C(u)=H_k(j_1u_{i_1},\ldots,j_Ku_{i_K})=1  ~.
\end{align*}
\proofbox

By a simple scaling argument we can prove corollary \ref{cor:dnf_small}.

\proof (of corollary \ref{cor:dnf_small})
By theorem \ref{thm:dnf_few_clauses}, it is $\srcsp$-hard to learn $\dnf^{n}$. Since $\dnf^{n}\subset \dnf_{n^2}$, we conclude that it is $\srcsp$-hard to learn $\dnf_{n^2}$. To establish the corollary, we note that $\dnf_{n^2}$ can be efficiently realized by $\dnf_{n^{\epsilon}}$ using the mapping $f:\{\pm 1\}^{n}\to \{\pm 1\}^{n^{\frac{2}{\epsilon}}}$ that pads the original $n$ coordinates with $n^{\frac{2}{\epsilon}}-n$ ones.
\proofbox

\subsection{Agnostically learning halfspaces}
\proof (of theorem \ref{thm:halfspaces})
Let $\ch$ be the hypothesis class of halfspaces over $\{-1,1,0\}^n$, induced by $\pm 1$ vectors.
We will show that agnostically learning $\ch$ is $\srcsp$-hard. While
we defined the class of $\half$ over instances in $\{\pm 1\}^n$,
proving the hardness of learning $\ch$ (which is defined over
$\{-1,1,0\}^n$) suffices for our needs, since $\ch$ can be efficiently
realized by $\half$ as follows: Define $\psi:\{-1,1,0\}\to \{\pm 1\}^2$ by
\[
\psi(\alpha)=
\begin{cases}
(-1,-1) & \alpha = -1
\\
(1,1) & \alpha = 1
\\
(-1,1) & \alpha = 0
\end{cases}~.
\]
Now define $\Psi:\{-1,1,0\}^n\to \{\pm 1\}^{2n}$ by
\[
\Psi(x)=(\psi(x_1),\ldots,\psi(x_n))~.
\]
Also define $\Phi:\{\pm 1\}^n\to \mathbb \{\pm 1\}^{2n}$ by
\[
\Phi(w)=(w_1,w_1,w_2,w_2,\ldots,w_n,w_n)~.
\]
It is not hard to see that for every $w\in \{\pm 1\}^n$ and every $x\in \{-1,1,0\}^n$, $h_w(x)=h_{\Phi(w)}(\Psi(x))$. Therefore, $\ch$ is efficiently realized by $\half$.

We will use assumption \ref{hyp:not_sat} with respect to the majority predicate $\maj_K:\{\pm 1\}^K\to \{0,1\}$. Recall that $\maj(x)=1$ if and only if $\sum_{i=1}^Kx_i>0$. The following claim analyses its relevant properties.
\begin{claim}\label{claim:maj_pred}
For every odd $K$,
\begin{itemize}
\item $\Uval(\maj_K) =1-\frac{1}{K+1}$.
\item $\var_0(\maj_K)=\frac{K+1}{2}$.
\end{itemize}
\end{claim}
\proof
It is clear that $\maj_K$ has $\frac{K+1}{2}$ $0$-variability.
We show next that $\Uval(\maj_K)=1-\frac{1}{K+1}$. Suppose that $K=2t+1$. Consider the distribution $\cd$ on $\{\pm 1\}^K$ defined as follows. With probability $\frac{1}{2t+2}$ choose the all zero vector, and with probability $\frac{2t+1}{2t+2}$ choose a vector uniformly at random among all vectors with $t+1$ ones. It is clear that $\E_{x\sim \cd}[\maj_K(x)]=1-\frac{1}{2t+2}$. We claim that $\cd$ is pairwise uniform, therefore, $\Uval(\maj_K)\ge 1-\frac{1}{2t+2}$. Indeed for every distinct $i,j\in [K]$,
\begin{align*}
&\Pr_{x\sim \cd}\left((x_i,x_j)=(0,1)\right)=\Pr_{x\sim \cd}\left((x_i,x_j)=(1,0)\right)=\frac{2t+1}{2t+2}\cdot
\frac{t+1}{2t+1}\cdot \frac{t}{2t}=\frac{1}{4} ~,  \\
&\Pr_{x\sim \cd}\left((x_i,x_j)=(1,1)\right)=\frac{2t+1}{2t+2}\cdot
\frac{t+1}{2t+1}\cdot \frac{t}{2t}=\frac{1}{4}~,
\end{align*}
and $\Pr_{x\sim \cd}\left((x_i,x_j)=(0,0)\right)=\frac{1}{4}$.

Next, we show that $\Uval(\maj_K)\le 1-\frac{1}{2t+t}$. Let $\cd$ be a pairwise uniform distribution on $\{\pm 1\}^K$. We have $\E_{x\sim \cd}\left[\sum_{i=1}^K\frac{x_i+1}{2}\right]=\frac{K}{2}$ therefore, by Markov's inequality,
\[
\E_{x\sim \cd}\left[\maj_K(x)\right]=\Pr_{x\sim \cd}\left(\maj_K(x)=1\right)=\Pr_{x\sim \cd}\left(\sum_{i=1}^K\frac{x_i+1}{2}\ge t+1\right)\le \frac{2t+1}{2(t+1)}~.
\]
Since this is true for every pairwise uniform distribution, $\Uval(\maj_K)\le \frac{2t+1}{2(t+1)}=1-\frac{1}{K+1}$.
\proofbox

Fix $\alpha\ge 1$. We will use theorem \ref{thm:basic_agnostic} to show that there is no efficient algorithm that approximately agnostically learns $\ch$ with approximation ratio of $\alpha$, unless the $\srcsp$ assumption is false.
Let $c>1$ and denote $\beta=\frac{1}{10\alpha}$. It suffices to show that there is a polynomial ensamble $\cd=\{\cd_n^{m(n)}\}_{n=1}^\infty$ that is $(\Omega(n^c),\alpha\beta+\frac{1}{n})$-scattered and it is $\srcsp$-hard to distinguish between a $\cd$-random sample and an $\beta$-almost realizable sample.

By assumption \ref{hyp:not_sat} and claim \ref{claim:maj_pred}, for large enough odd $K$, it is $\srcsp$-hard to distinguish between a random instances of $\csp(\maj_K)$ with $m(n)= n^c$ constraints and instances with value $\ge 1-\beta$. Consider the following ensamble $\cd=\{\cd_n^{2m(n)}\}_{n=1}^\infty$: pick $m=m(n)$ independent uniform vectors
$x_1,\ldots,x_m\in\{x\in\{-1,1,0\}^n\mid |\{i\mid x_i\ne 0\}|=K|\}$. Then, consider the sample
$S=\{(x_1,1),(-x_1,0),\ldots,(x_m,1),(-x_m,0)\}$. The theorem follows from the following claim:
\begin{claim}~
\begin{itemize}
\item $\cd$ is $(\Omega(n^c),\alpha\beta+\frac{1}{n})$-scattered.
\item It is $\srcsp$-hard to distinguish between a $\cd$-random sample and an $\beta$-almost realizable sample.
\end{itemize}
\end{claim}
\proof
The first part follows from proposition \ref{prop:scat_crit}. Next, we show that it is $\srcsp$-hard to distinguish between a $\cd$-random sample and $\beta$-almost realizable sample. We will reduce from the problem of distinguishing between a random instance with $m(n)$ constraints and an instance with value $\ge 1-\beta$. Given an 
instance $J$ with $m(n)$ constraints, we will produce a sample $S$ of $2m(n)$ examples by transforming each constraint into two examples 
as follows: for the constraint $C(x)=\maj(j_1x_{i_1},\ldots,{j_K}x_{i_K})$
we denote by $u(C)\in \{x\in\{-1,1,0\}^n\mid |\{i\mid x_i\ne 0\}|=K|\}$ the vector whose $i_{l}$ coordinate is $j_l$. We will produce the examples $(u(C),1)$ and $(-u(C),0)$. It is not hard to see that if 
$J$ is random then $S\sim  \cd^{2m(n)}_n$. If the value of $J$ is $\ge 1-\beta$, indicated by an assignment $w\in \{\pm 1\}^n$, it is not hard to see 
that $h_w\in\ch$ $\epsilon$-almost realizes the sample $S$. This concludes the proof of the claim.
\proofbox

Combining all the above we conclude the proof of theorem \ref{thm:halfspaces}.
\proofbox

\begin{remark}\label{rem:halfsapces}
What is the real approximation ratio of agnostically learning
halfspaces in $n$ dimension? Taking a close look at the above proof,
we see that in some sense, by the $\srcsp$ assumption with $\maj_{K}$,
it is hard to agnostically  learn halfspaces with approximation ratio
of $\Omega\left(K\right)$. If we let $K$ grow with $n$ (this is {\em
  not} allowed by the $\srcsp$-hypothesis), say $K=\frac{1}{100}n$, we can hypothesize that it is hard to agnostically  learn halfspaces with approximation ratio of about $n$. The approximation ratio of the best known algorithms is somewhat better, namely, $\frac{n}{\log(n)}$. But this is not very far from our guess. Therefore, one might hypothesize that the best possible approximation ratio is, say, of the form $\frac{n}{\poly\left(\log(n)\right)}$. Given a rigorous treatment to the above intuition is left as an open question.
\end{remark}

\begin{remark}\label{rem:large_margin}
The problem of {\em learning large margin halfsapces} is an important variant of the problem of learning halfspaces. Here, we assume that the instance space is the unit ball in $\mathbb R^d$. For $1>\gamma>0$, the $\gamma$-margin error of a hyperplane $h$ is the probability of an
example to fall on the wrong side of $h$ or at a distance $\le\gamma$
from it. The $\gamma$-margin error of the best $h$ (with respect to a distribution $\cd$) is denoted
$\Err_\gamma(\D)$. An $\alpha(\gamma)$-approximation algorithm
receives $\gamma,\epsilon$ as input and outputs a classifier with error rate $\le
\alpha(\gamma)\Err_\gamma(\D) + \epsilon$.  Such an algorithm is
efficient if it uses $\poly(\frac{1}{\gamma},\frac{1}{\epsilon})$
samples and runs in time polynomial in the sample size. For a detailed definition, the reader is referred to \cite{danielyLinSha13The}.

It is not hard to see that the proof of theorem \ref{thm:halfspaces} shows that it is hard to approximately learn large margin halfspaces with any constant approximation ratio. Taking considerations as in remark \ref{rem:halfsapces}, one might hypothesize that the correct approximation ratio for this problem is about $\frac{1}{\gamma}$. As 
in the case of learning halfspaces, best known algorithms \cite{LongSe11, BirnbaumSh12} do just a bit better, namely, they have an approximation ratio of $\frac{1}{\gamma\sqrt{\log(1/\gamma)}}$. 
Therefore, one might hypothesize that the best possible approximation ratio is $\frac{1}{\gamma\poly\left(\log(1/\gamma)\right)}$. We note that a recent result \cite{danielyLinSha13The} shows that this is the best possible 
approximation ratio, if we restrict ourselves to a large class of learning algorithms (that includes SVM with a kernel, regression, Fourier transform and more).
\end{remark}

\subsection{Learning automata}
For a function $q:\mathbb N\to\mathbb N$, let $\auto_{q(n)}$ be the
class of functions $h:\{\pm 1\}^n\to \{0,1\}$ that can be realized by
a finite automaton with $q(n)$ states.
\begin{theorem}\label{thm:auto}
For every $\epsilon>0$, it is $\srcsp$-hard to learn $\auto_{n^\epsilon}$.
\end{theorem}
\begin{note}
The theorem remains true (with the same proof), even if we restrict to acyclic automata.
\end{note}
\proof
By a simple scaling argument, as in the proof of corollary
\ref{cor:dnf_small}, it is enough to show that it is $\srcsp$-hard to learn $\auto_{n^2+1}$. By theorem \ref{thm:dnf_few_clauses}, it is $\srcsp$-hard to learn $\dnf^{\log_2(n)}$. To establish the theorem, we will show that if a function $h:\{\pm 1\}^n\to\{0,1\}$ can be realized by a $\dnf$ formula with $\log_2(n)$ clauses, then it can be realized by an automaton with $n^2+1$ states.

For simplicity, assume that $n$ is a power of $2$. Given a $\dnf$
formula $R$ with $k:=\log_2(n)$ clauses, we will construct an acyclic
automaton as follows. For each variable we will have $n$ states
(corresponding to subsets of $[k]$). In addition, we will have a start state. From the start state, the automaton will jump to the state $(1,A)$, where $A$ is the set of the indices of all the clauses in $R$ that are not violated by the 
value of $x_1$. After reading $x_2$ the automaton will jump to the state $(2,A)$, where $A$ is the set of the indices of all the clauses in $R$ that are not violated by the values of $x_1$ and $x_2$. In this manner, after reading $x_1,\ldots, x_n$ the automaton will be at the state $(n,A)$, where $A$ is the set of the indices of all the clauses in $R$ that are satisfied by $x_1,\ldots,x_n$. The automaton accepts if and only if $A\ne \emptyset$.

Clearly, this automaton calculates the same function as $R$.
\proofbox

\subsection{Toward intersection of $4$ halfspaces}\label{sec:4half}
For $1\le l\le k$ Consider the predicate $T_{k,l}:\{\pm 1\}^k\to \{0,1\}$ such that $T_{k,l}(x)=1$ if and only if $x$ has at least $l$ ones. For example, $T_{k,1}$ is the $\mathrm{SAT}$ predicate, $T_{k,\lfloor \frac{k}{2}\rfloor+1}$ is the $\mathrm{MAJ}$ predicate and $T_{k,k}$ is the $\mathrm{AND}$ predicate.
Define $P_{k}:\left(\{\pm 1\}^{k}\right)^{8}\to\{0,1\}$ by
\begin{eqnarray*}
P_{K}(x^1,\ldots,x^8)&=&\left(\wedge_{j=1}^4 T_{k,\lceil \frac{k}{2}\rceil-1}(x^j)\right)
\wedge
\neg
\left(\wedge_{j=5}^8 T_{k,\lceil \frac{k}{2}\rceil-1}(x^j)\right)~.
\end{eqnarray*}

\begin{proposition}
There is $k_0$ such that for every odd $k\ge k_0$ we have
\begin{enumerate}
\item Assuming the unique games conjecture, $P_k$ is heredity approximation resistant.
\item For some constant $1>\alpha>0$, it is $\mathbf{NP}$-hard to distinguish between satisfiable instances to $\csp(P_k)$ and instances with value $\le \alpha$.
\end{enumerate}
\end{proposition}
\proof
We start with part 1. By \citep{austrin2009approximation}, it suffices to show that there is a pairwise uniform distribution that is supported in $P_k^{-1}(1)$. Denote $Q(x^1,\ldots x^4)=\wedge_{j=1}^4 T_{k,\lceil \frac{k}{2}\rceil-1}(x^j)$ and $R(x^1,\ldots x^4)=\neg\left(\wedge_{j=1}^4 T_{k,\lceil \frac{k}{2}\rceil-1}(x^j)\right)$. Note that if $\cd_Q$ is a pairwise uniform distribution that is supported in $Q^{-1}(1)$ and $\cd_R$ is a pairwise uniform distribution that is supported in $R^{-1}(1)$, then $\cd_{Q}\times \cd_R$ is a pairwise uniform distribution that is supported in $P_k^{-1}(1)$. Therefore, it suffices to show that such $\cd_Q$ and $\cd_R$ exist.

We first construct $\cd_{Q}$. Let $\cd_k$ be the following
distribution over $\{\pm 1\}^k$ -- with probability $\frac{1}{k+1}$
choose the all-one vector and with probability $\frac{k}{k+1}$, choose
at random a vector with $\lceil\frac{k}{2}\rceil-1$ ones (uniformly
among all such vectors). By the argument of claim \ref{claim:maj_pred}, $\cd_k$ is pairwise uniform. Clearly, the distribution $\cd_Q=\cd_k\times\cd_k\times\cd_k\times\cd_k$ over $\left(\{\pm 1\}^k\right)^4$ is a pairwise uniform distribution that is supported in $Q^{-1}(1)$.

Next, we construct $\cd_{R}$. Let $k_0$ be large enough so that for every $k\ge k_0$, the probability that a random vector from $\{\pm 1\}^k$ will have more than $\lceil \frac{k}{2}\rceil$ minus-ones is $\ge \frac{3}{8}$ (it is easy to see that this probability approaches $\frac{1}{2}$ as $k$ approaches $\infty$. Therefore, such $k_0$ exists). Now, let $Z\in \{0,1\}^4$ be a random variable that satisfies:
\begin{itemize}
\item $Z_1,\ldots,Z_4$ are pairwise independent.
\item For every $1\le i\le 4$, $\Pr(Z_i=1)=\frac{3}{8}$.
\item $\Pr(Z=(0,0,0,0))=0$.
\end{itemize}
In a moment, we will show that a random variable with the above
properties exists. Now, let $B\subset \{\pm 1\}^k$ be a set with $|B|\ge \frac{3}{8}\cdot 2^k$ such that every vector in $B$ has more than $\lceil\frac{k}{2}\rceil$ minus-ones. Consider the distribution $\cd_R$ of the random variable $(X^1,\ldots,X^4)\in \left(\{\pm 1\}^k\right)^4$ sampled as follows. We first sample $Z$, then, for $1\le i\le 4$, if $Z_i=1$, we choose $X^i$ to be a random vector $B$ and otherwise, we choose $X^i$ to be a random vector $B^c$.

We note that since $Z_1,\ldots,Z_4$ are pairwise independent, $X^1,\ldots, X^4$ are pairwise independent as well. Also, the distribution of $X^i,\;1=1,\ldots,4$ is uniform. Therefore, $\cd_R$ is pairwise uniform. Also, since $\Pr(Z=(0,0,0,0))=0$, with probability $1$, at least one of the $X^i$'s will have more than $\lceil\frac{k}{2}\rceil$ minus-ones. Therefore, $\cd_R$ is supported in $R^{-1}(1)$.

It is left to show that there exists a random variable $Z\in \{0,1\}^4$ as specified above. Let $Z$ be the random variable defined as follows:
\begin{itemize}
\item With probability $\frac{140}{192}$ $Z$ is a uniform vector with a single positive coordinate.
\item With probability $\frac{30}{192}$ $Z$ is a uniform vector with $2$ positive coordinates.
\item With probability $\frac{22}{192}$ $Z$ is a uniform vector with $4$ positive coordinates.
\end{itemize}
Clearly, $\Pr(Z=(0,0,0,0))=0$. Also, for every distinct $1\le i,j\le 4$ we have
\[
\Pr(Z_i=1)=\frac{140}{192}\cdot\frac{1}{4}+\frac{30}{192}\cdot\frac{1}{2}+
\frac{22}{192}=\frac{3}{8}
\]
and
\[
\Pr(Z_i=1,Z_j=1)=\frac{30}{192}\cdot\frac{1}{6}+
\frac{22}{192}=\left(\frac{3}{8}\right)^2~.
\]
Therefore, the other two specifications of $Z$ hold as well.

We proceed to part 2. The reduction is quite simple and we only sketch it. By adding dummy variables, it is enough to prove that it is $\np$-hard to distinguish between satisfiable instances of $\csp(T_{k,\lceil\frac{k}{2}\rceil-1})$ and instances with value $\le \alpha$ for some constant $0<\alpha<1$. We will show somewhat stronger property, namely, that if $1\le l\le k-2$, then for some $0<\alpha<1$, it is $\np$-hard to distinguish between satisfiable instances of $\csp(T_{k,l})$ and instances with value $\le \alpha$.

We will reduce from the problem of distinguishing between satisfiable
instances to $3$-$\sat$ and instances with value $\le
\frac{8}{9}$. This problem is $\np$-hard \citep{haastad2001some}. Given an instance $J$ to $3$-$\sat$, we will produce an instance $R(J)$ to $\csp(T_{k,l})$ as follows. Its variables would be the variables of $J$ together with some new variables. For every constraint $C(x)=j_1x_{i_1}\vee j_2x_{i_2}\vee j_3x_{i_3}$ in $J$, we will add $k+l-1$ new variables $x^C_1,\ldots,x^C_k$ and $y^C_4,\ldots,x^C_{l+2}$. These new variables will be used only in the new clauses corresponding to $C$. We will introduce the following constraints: we add the constraint $T_{k,l}(j_1x_{i_1},j_2x_{i_2},j_3x_{i_3},y^C_4,\ldots,y^C_{l+2},-x^C_{l+3},\ldots,-x^C_k)$. Also, for every $(j_1,\ldots,j_k)\in \{\pm 1\}^k$ with at most $(k-l)$ minus-ones we will add the constraint $T_{k,l}(j_1x^C_1,\ldots,j_kx^C_k)$. 

If $J$ is satisfiable, then $R(J)$ is satisfiable as well: simply set all new variables to $1$. On the other hand, if $\val (J)\le\frac{8}{9}$, then it is not hard to see that for every 
assignment to $R(J)$'s variables, for at least $\frac{1}{9}$ of $J$'s clauses, at least one of the new clauses corresponding to it will be unsatisfied. Since we introduce $\le 2^k$ constraints in $R(J)$ for each constraint in $J$, we conclude that $\val(R(J))\le 1-2^{-k}\frac{1}{9}$. Therefore, the theorem holds with $\alpha=1-2^{-k}\frac{1}{9}$. 
\proofbox

\begin{conjecture}\label{conj:inter_of_two}
$P_k$ is heredity approximation resistant on satisfiable instances.
\end{conjecture}
\begin{theorem}\label{thm:inter_of_two}
Assuming conjecture \ref{conj:inter_of_two}, it is $\srcsp$-hard to learn $\inter_{4}$
\end{theorem}
\proof (sketch)
The proof goes along the same lines of the proof of theorem \ref{thm:halfspaces}. We will prove $\srcsp$-hardness for learning intersections of two halfspaces over $\{-1,1,0\}^n$, induced by $\pm 1$ vectors. As in the proof of theorem \ref{thm:halfspaces}, $\srcsp$-hardness of learning intersections of two halfspaces over the boolean cube follows from this.

Fix $d>0$. It is not hard to check that $\var_0(P_k)\ge \lceil\frac{k}{2}\rceil-2$. Therefore, by conjecture \ref{conj:inter_of_two} and assumption \ref{hyp:sat}, for large enough odd $k$, it is $\srcsp$-hard to distinguish between a random instance to $\csp(P_k)$ with $n^d$ constraints and a satisfiable instance. We will reduce from this problem to the problem of distinguishing between a realizable sample and a random sample that is $(\Omega(m^d),\frac{1}{5})$-scattered. Since $d$ is arbitrary, the theorem follows.

Given an instance $J$, we produce two examples for each constraint: for the constraint

\begin{eqnarray*}
C(x) &=&\left(\wedge_{q=1}^4 T_{k,\lceil \frac{k}{2}\rceil-1}(j_{q,1}x_{i_{q,1}},\ldots,j_{q,k}x_{i_{q,k}})\right)
\\
&& \;\;\;\;\;\;\;\;\;\wedge
\neg \left(\wedge_{q=5}^8 T_{k,\lceil \frac{k}{2}\rceil-1}(j_{q,1}x_{i_{q,1}},\ldots,j_{q,k}x_{i_{q,k}})\right)
\end{eqnarray*}
we will produce two examples in $\{-1,1,0\}^{4n}\times \{0,1\}$, each of which has exactly $4k$ non zero coordinates. The first is a positively labelled example whose instance is the vector with the value $j_{q,l},\;1\le q\le 4,1\le l\le k$ in the $n(q-1)+i_{q,l}$ coordinate. 
the second is a negatively labelled example whose instance is the vector with the value $j_{q,l},\;5\le q\le 8,1\le l\le k$ in the $n(q-5)+i_{q,l}$ coordinate. 

It is not hard to see that if $J$ is satisfiable then the produced sample is realizable by intersection of four halfspaces: if $u\in \{\pm 1\}^n$ is a satisfying assignment then the sample is realized by the intersection of the $4$ halfspaces $\sum_{i=1}^{n} u_ix_{n(q-1)+i}\ge -1,\;\;q=1,2,3,4$. On the other hand, by proposition \ref{prop:scat_crit}, if $J$ is random instance with $n^d$ constraints, then the resulting ensamble is $(\Omega(n^d),\frac{1}{5})$ scattered. 

\proofbox

\subsection{Agnostically learning parity}
For convenience, in this section the domain of hypotheses will be
$\{0,1\}^n$  and the domain of predicates will be $\{0,1\}^K$ (instead of $\{\pm 1\}^n$ and $\{\pm 1\}^K$). For every $S\subset [n]$ define $\chi_S:\{0, 1\}^n\to \{\pm 1\}$ by $\chi_S(x)=\oplus_{i\in S}x_i$. Let $\parity$ be the hypothesis class consisting of all functions $\chi_S,\;S\subset [n]$.
\begin{theorem}\label{thm:parity}
For every constant $\alpha\ge 1$, it is $\srcsp$-hard to approximately agnostically learn $\parity$ with an approximation ratio of $\alpha$.
\end{theorem}
\proof 
Let $P_K:\{0, 1\}^K\to \{0,1\}$ be the parity predicate. That is, $P_K(x)=\oplus_{i=1}^Kx_i$. 
We first show that for $K\ge 3$, $\Uval(P_K)=1$. Indeed, a pairwise uniform distribution which is supported in $P_K^{-1}(1)$ is the following -- choose $(x_1,\ldots,x_{K-1})$ uniformly at random and then choose $x_{K}$ so that $P_K(x)=1$. Second, it is clear that $\var_0(P_K)=K$. Therefore, by assumption \ref{hyp:not_sat}, for every $\beta > 0$ and every $d$, for sufficiently large $K$, it is $\srcsp$-hard to distinguish between instances to $\csp(P_K)$ with value $\ge 1-\beta$ and random instances with $m^d$ constraints. Note that with the convention that the domain of $P_K$ is $\{0,1\}^K$, the constraints of instances to $\csp(P_K)$ are of the form $C(x)=x_{i_1}\oplus\ldots \oplus x_{i_K}$ or $C(x)=x_{i_1}\oplus\ldots \oplus x_{i_K}\oplus 1$.

We will reduce from the aforementioned problem to the problem of distinguishing between $\beta$-almost realizable sample and $\cd$-random sample for a distribution $\cd$ which is $\left(\Omega\left(n^d\right),\frac{1}{4}\right)$-scattered. Since both $\beta$ and $d$ are arbitrary, the theorem follows from theorem \ref{thm:basic_agnostic}.

Given an instance $J$ to $\csp(P_K)$, for each constraint
$C(x)=x_{i_1}\oplus\ldots \oplus x_{i_K}\oplus b$ we will generate an
example $(u_C,y_C)$ where $u_C$ is the vector with ones precisely in
the coordinates $i_1,\ldots,i_K$ and $y_C=b$. It is not hard to verify
that if $J$ is a random instance with $m^d$ constraints then the
generated sample is $\left(\Omega\left(n^d\right),\frac{1}{4}\right)$-scattered. On the other hand, assume that the assignment $\psi\in \{0, 1\}^n$ satisfies $1-\beta$ fraction of the constraints. Consider the hypothesis $\chi_S$ where $S=\{i\mid x_i=1\}$. We have $\chi_S(x_C)=\oplus_{i\in S}(u_C)_i=\oplus_{q=1}^K\psi_{i_q}$. Therefore, $\psi$ satisfies $C$ if and only if $\chi_S$ is correct on $(u_C,y_C)$. Since $\psi$ satisfies $1-\beta$ fraction of the constraints, the generated sample is $\beta$-almost realizable.
\proofbox

\section{Resolution lower bounds}\label{sec:res}
In this section we prove theorem \ref{thm:res_main}. Let $P:\{0,1\}^K\to\{0,1\}$ be some predicate. Let $\tau=\{T_1,\ldots, T_r\}$ be a resolution refutation for a $\csp(P)$ instance $J$.
A basic parameter associated with $\tau$ is the {\em width}. The \emph{width} of a clause is the number of literals it contains, and the {\em width} of $\tau$ is $\width(\tau):=\max_{1\le i\le r}\width(T_i)$. We also define the width of an unsatisfiable instance $J$ to $\csp(P)$ as the minimal width of a resolution refutation of $J$.
Ben-Sasson and Wigderson \cite{BenWi99} have shown that if an instance to $\csp(P)$ has a short resolution refutation, then it necessarily has a narrow resolution refutation. Namely,
\begin{theorem}[\cite{BenWi99}]\label{thm:ben_sasson_wig}
Let $J$ be an unsatisfiable instance to $\csp(P)$. The length of every resolution refutation for $J$ is at least $2^{\Omega\left(\frac{\width^2(J)}{n}\right)}$.
\end{theorem}
Theorem \ref{thm:res_main} now follows from theorem \ref{thm:ben_sasson_wig} and the following two lemmas.
\begin{lemma}\label{lem:width_criterion}
Let $J$ be an unsatisfiable instance to $\csp(P)$. Assume that for every subset $I$ of $l$ constraints from $J$, most of the constraints in $I$ have $\ge K-\var_0(P)-1$ variables that do not appear in any other constraint in $I$. Then $\width(J)\ge \frac{l}{6}$.
\end{lemma}
\proof
Let $\tau=\{T_1,\ldots,T_r\}$ be a resolution refutation to $J$. Define $\mu(T_i)$ as the minimal number $\mu$ such that $T_i$ is implied by $\mu$ constraints in $J$.
\begin{claim}~
\begin{enumerate}
\item $\mu(\emptyset)> l$.
\item If $T_i$ is implied by $T_{i_1},T_{i_2},\;i_1,i_2<i$ then $\mu(T_i)\le \mu(T_{i_1})+\mu(T_{i_2})$.
\end{enumerate}
\end{claim}
\proof
The second property clearly holds. To prove the first property,
suppose toward a contradiction that $\mu(\emptyset)\le l$. It follows
that there are $t\le l$ constraints $I\subset J$ that implies the
empty clause, i.e., it is impossible to simultaneously satisfy all the
constraints in $I$. By the assumption of the lemma, it is possible to
choose an ordering $I=\{C_1,\ldots,C_t\}$ such that for every $1\le
i\le t$, $C_i$ contains at least $K-\var_0(P)-1$ variables that do not
appear in $C_1,\ldots,C_{i-1}$. Indeed, let us simply take $C_t$ to be
a clause that contains at least $K-\var_0(P)-1$ variables that do not
appear in the clauses in $I\setminus\{C_t\}$. Then, choose $C_{t-1}$
in the same way from $I\setminus\{C_t\}$ and so on. Now, let
$\psi\in\{\pm 1\}^n$ be an arbitrary assignment that satisfies
$C_1$. By the definition of $0$-variability, it is possible to change
the values of the variables appearing in $C_2$ but not in $C_1$ to
satisfy also $C_2$. We can continue doing so till we reach an
assignment that satisfies $C_1,\ldots, C_t$ simultaneously. This leads
to the desired contradiction.
\proofbox

By the claim, and the fact that $\mu(C)=1$ for every clause that is implied by one of the constraints of $J$, we conclude that there is some $T_i$ with $\frac{l}{3}\le \mu=\mu(T_j)\le \frac{2l}{3}$. It follows that there are $\mu$ constraints $C_1,\ldots, C_\mu$ in $J$ that imply $T_j$, but no strict subset of these clauses implies $T_j$. For simplicity, assume that these constraints are ordered such that for every $1\le i\le \frac{\mu}{2}$, $C_i$ contains at least $K-\var_0(P)-1$ variables that do not appear in the rest of these constraints. The proof of the lemma is established by the following claim
\begin{claim}
For every $1\le i\le \frac{\mu}{2}$, $T_j$ contains a variable appearing only is $C_i$.
\end{claim}
\proof
Assume toward a contradiction that the claim does not hold for some
$1\le i\le \frac{\mu}{2}$. Since no strict subset of
$C_1,\ldots,C_\mu$ imply $T_j$, there is an assignment $\psi\in\{\pm
1\}^n$ such that for every $i'\ne i$, $C_{i'}(\psi)=1$ but
$T_j(\psi)=0$. Since $C_1,\ldots,C_\mu$ imply $T_j$, we must have
$C_i(\psi)=0$. Now, by the definition of $0$-variability, we can
modify the values of the $K-\var_0(P)-1$ variables that appear only in $C_i$ to have a new assignment $\psi'\in\{\pm 1\}^n$ with $C_i(\psi')=1$. Since $T_j$ and the rest of the constraints do not contain these variables, we conclude that still for every $i'\ne i$, $C_{i'}(\psi)=1$ and $T_j(\psi)=0$. This contradicts the fact that $C_1,\ldots,C_\mu$ imply $T_j$.
\proofbox

\proofbox

The next lemma shows that the condition in lemma \ref{lem:width_criterion} holds w.h.p. for a suitable random instance. For the sake of readability, it is formulated in terms of sets instead of constraints.

\begin{lemma}
Fix integers $k>r>d$ such that $r>\max\{17d,544\}$. Suppose that $A_1,\ldots, A_{n^d}\in \binom{[n]}{k}$ are chosen uniformly at random. Then, with probability $1-o_n(1)$, for every $I\subset [n^d]$ with $|I|\le n^{\frac{3}{4}}$ for most $i\in I$ we have $|A_i\setminus \cup_{j\in I\setminus \{i\}}A_j|\ge k-r$.
\end{lemma}
\proof
Fix a set $I$ with $2\le t\le n^{\frac{3}{4}}$ elements. Order the sets in $I$ arbitrarily and also order the elements in each set arbitrarily. Let $X_1,\ldots,X_{kt}$ be the following random variables: $X_1$ is the first element in the first set of $I$, $X_2$ is the second element in the first set of $I$ and so on till the $k$'th element of the last set of $I$.

Denote by $R_i\;\;1\le i\le kt$ the indicator random variable of the
event that $X_i=X_j$ for some $j<i$. We claim that if $\sum R_i<
\frac{tr}{4}$, the conclusion of the lemma holds for $I$. Indeed, let
$J_1\subset I$ be the set of indices with $R_i=1$, $J_2\subset I$ be
the set of indices $i$ with $R_i=0$ but $X_i=X_j$ for some $j>i$ and
$J=J_1\cup J_2$. If the conclusion of the lemma does not hold for $I$, then $|J|\ge \frac{tr}{2}$. If in addition $|J_1|=\sum R_i< \frac{tr}{4}$ we must have $|J_2|>\frac{tr}{4}>|J_1|$. For every $i\in J_2$, let $f(i)$ be the minimal index $j>i$ such that $X_i=X_j$. We note that $f(i)\in J_1$, therefore $f$ is a mapping from $J_2$ to $J_1$. Since $|J_2|>|J_1|$, $f(i_1)=f(i_2)$ for some $i_1<i_2$ in $J_2$. Therefore, $X_{i_1}=X_{f(i_1)}=X_{i_2}$ and hence, $R_{i_2}=1$ contradicting the assumption that $i_2\in J_2$.

Note that the probability that $R_i=1$ is at most $\frac{tk}{n}$. This estimate holds also given the values of $R_1,\ldots, R_{i-1}$. It follows that the probability that $R_i=1$ for every $i\in A$ for a particular $A\subset I$ with $|A|=\lceil\frac{rt}{4}\rceil$ is at most $\left(\frac{tk}{n}\right)^{\frac{rt}{4}}$. Therefore, for some constants $C',C>0$ (that depend only on $d$ and $k$), the probability that $J$ fails to satisfy the conclusion of the lemma is bounded by
\begin{eqnarray*}
\Pr\left(\sum R_i\ge \frac{tr}{4}\right) &\le & \binom{tk}{\lceil\frac{tr}{4}\rceil}\left(\frac{tk}{n}\right)^{\frac{tr}{4}}
\\
&\le & 2^{C\cdot t}\left(\frac{tk}{n}\right)^{\frac{tr}{4}}
\\
&\le & 2^{C'\cdot t}\left(\frac{t}{n}\right)^{\frac{tr}{4}}
\end{eqnarray*}
The second inequality follows from Stirling's approximation. Summing over all collections $I$ of size $t$ we conclude that for some $C''>0$, the probability that the conclusion of the lemma does not hold for some collection of size $t$ is at most
\[
\binom{n^d}{t}2^{C'\cdot t}\left(\frac{t}{n}\right)^{\frac{tr}{4}}
\le 
n^{dt-\frac{1}{16}tr}\cdot2^{C'\cdot t}
\le 
n^{-\frac{1}{272}tr}\cdot2^{C'\cdot t}
\le 
n^{-2t}\cdot2^{C'\cdot t}\le
C''\frac{1}{n}
\]
Summing over all $2\le t\le n^{\frac{3}{4}}$, we conclude that the probability that the conclusion of the lemma does not hold  is at most $C'' n^{-\frac{1}{4}}=o_n(1)$.
\proofbox

\section{On basing the $\srcsp$ assumption on $\mathbf{NP}$-Hardness}\label{sec:base_on_np}
Fix a predicate $P:\{\pm 1\}^{K}\to \{0,1\}$ and let $1\ge \alpha >\Lval(P)$. Let $L\subset \{0,1\}^*$ be some language. We say that $L$ can be efficiently reduced to the problem of distinguishing between random instances to $\csp(P)$ with $Cn$ constraints and instances with value $\ge \alpha$, if there is an efficient probabilistic Turing machine that given $x\in \{0,1\}^n$, acts as follows: for some function $f:\mathbb N\to\mathbb N$,
\begin{itemize}
\item If $x\in L$ then $M(x)$ is an instance to $\csp(P)$ with $f(n)$ variables, $C\cdot f(n)$ constraints and value $\ge \alpha$.
\item If $x\notin L$ then $M(x)$ is a random instance to $\csp(P)$ with $f(n)$ variables and $C\cdot f(n)$ constraints.
\end{itemize}

\begin{theorem}\label{thm:base_on_np}
For every sufficiently large constant $C>0$, the following holds.
Assume that the language $L\subset \{0,1\}^*$ can be efficiently reduced to the problem of distinguishing between random instances to $\csp(P)$ with $m(n)\ge Cn$ constraints and instances with value $\ge \alpha$. Then, $L$ has a statistical zero knowledge proof.
\end{theorem}

\begin{corollary}\label{cor:base_on_np}
For every sufficiently large constant $C>0$, the following holds. Assume that there is a reduction from an either an $\mathbf{NP}$-hard or $\mathbf{CoNP}$-hard problem to the problem of distinguishing between random instances to $\csp(P)$ with $m(n)\ge Cn$ constraints and instances with value $\ge \alpha$. Then, the polynomial hierarchy collapses.
\end{corollary}
\proof (of corollary \ref{cor:base_on_np})
Under the conditions of the corollary, by theorem \ref{thm:base_on_np}, we have $\mathbf{NP}\subset \mathbf{SZKP}$ or $\mathbf{CoNP}\subset \mathbf{SZKP}$. Since $\mathbf{SZKP}$ is closed under taking complement \citep{okamoto1996relationships}, in both cases, $\mathbf{NP}\subset \mathbf{SZKP}$. Since $\mathbf{SZKP}\subset \mathbf{CoAM}$ \citep{aiello1991statistical}, we conclude that $\mathbf{NP}\subset \mathbf{CoAM}$, which collapses the polynomial hierarchy \citep{bogdanov2006worst}.
\proofbox

\proof (of theorem \ref{thm:base_on_np})
Let $C>0$ be a constant large enough so that, with probability $\ge \frac{1}{2}$, a random instance to $\csp(P)$ with $Cn$ constraints will have value $\le \alpha$.

Consider the following problem. The input is a circuit $\Psi:\{0,1\}^n\to \{0,1\}^m$ and a number $t$. The instance is a YES instance if the entropy\footnote{We consider the standard Shannon's entropy with bits units.} of  $\Psi$, when it acts on a uniform input sampled from $\{0,1\}^n$, is $\le t-1$. The instance is a NO instance if this entropy is $\ge t$. By \citep{goldreich1999comparing} this problem is in $\mathbf{SZKP}$. To establish the proof, we will show that $L$ can be reduced to this problem.

Assume that there is a reduction from the language $L$ to the problem of distinguishing between random instances to $\csp(P)$ with $m(n)\ge Cn$ constraints and instances with value $\ge \alpha$. 
Let $M$ and $f$ be a Turing machine and a function that indicate that.
By a standard argument, it follows that there is an efficient deterministic Turing machine $M'$ that given $x\in \{0,1\}^n$ produces a circuit $\Psi$ whose input is $\{0,1\}^{g(n)}$ for some polynomially growing function and whose output is an instance to $\csp(P)$, such that, for a uniformly randomly chosen input $z\in \{0,1\}^{g(n)}$,
\begin{itemize}
\item If $x\in L$ then $\Psi(z)$ is a (possibly random) satisfiable instance to $\csp(P)$ with  $f(n)$ variables and $m(f(n))$ constraints.
\item If $x\notin L$ then $\Psi(z)$ is a random instance to $\csp(P)$ with  $f(n)$ variables and $m(f(n))$ constraints. 
\end{itemize}
Since the number of instances to $\csp(P)$ with $m(f(n))$ constraints is $\left(\binom{f(n)}{K}2^K\right)^{m(f(n))}$, in the second case, the entropy of $\Psi$ is $q(n):=m(f(n))\log_2\left(\binom{f(n)}{K}2^K\right)$. On the other hand, in the first case, the entropy is at most the entropy of a random instance to $\csp(P)$ with $m(f(n))$ constraints and value $\ge \alpha$. By the choice of $C$, the number of such instances is at most half of the total number of instances with $m(f(n))$ constraints. Therefore, the entropy of $\Psi$ is at most $m(f(n))\log_2\left(\binom{f(n)}{K}2^K\right)-1=q(n)-1$.
Hence, using $M'$, we can reduce $L$ to the problem mentioned in the beginning of the proof.
\proofbox

\paragraph{Acknowledgements:}
Amit Daniely is a recipient of the Google Europe Fellowship in Learning Theory, and this research is supported in part by this Google Fellowship. Nati Linial is supported by grants from ISF, BSF and I-Core. Shai Shalev-Shwartz is supported by the Israeli Science Foundation grant number 590-10. We thank Sangxia Huang for his kind help and for valuable discussions about his paper \citep{huang2013approximation}.
We thank Guy Kindler for valuable discussions.

\bibliography{bib}

\end{document}